\documentclass{article}
\usepackage{iclr2027_conference,times}

\providecommand{\paperversion}{preprint}
\def\submissionversion{submission}
\def\preprintversion{preprint}
\def\camerareadyversion{camera-ready}
\edef\paperversionexpanded{\paperversion}

\newif\ificlrpreprint
\iclrpreprintfalse
\ifx\paperversionexpanded\preprintversion
  \iclrpreprinttrue
  \iclrfinalcopy
\else\ifx\paperversionexpanded\camerareadyversion
  \iclrfinalcopy
\else\ifx\paperversionexpanded\submissionversion
\else
  \PackageError{main}{Unknown paper version `\paperversion'}%
    {Use submission, preprint, or camera-ready.}
\fi\fi\fi

\usepackage{amsmath,amsfonts,bm}

\def\eqref#1{equation~\ref{#1}}

\def\1{\bm{1}}

\DeclareMathAlphabet{\mathsfit}{\encodingdefault}{\sfdefault}{m}{sl}
\SetMathAlphabet{\mathsfit}{bold}{\encodingdefault}{\sfdefault}{bx}{n}

\usepackage{url}
\usepackage{graphicx}
\usepackage{wrapfig}
\usepackage{capt-of}
\usepackage{amsmath}
\usepackage{amsfonts}
\usepackage{booktabs}
\usepackage{latexsym}
\usepackage{multirow}
\usepackage{pifont}
\usepackage{algorithm}
\usepackage{algorithmic}
\usepackage[most]{tcolorbox}
\usepackage{hyperref}
\usepackage[table]{xcolor}

\definecolor{lightblue}{HTML}{4AA3DF}
\definecolor{titlegray}{HTML}{BFBFBF}
\definecolor{boxbg}{HTML}{F7F7F7}

\newcommand{\bfunderline}[1]{\textbf{\underline{#1}}}

\newtcolorbox{PromptStyleBox}[1]{
  enhanced,
  breakable,
  unbreakable,
  colback=boxbg,
  colframe=titlegray,
  boxrule=0.6pt,
  arc=2pt,
  left=10pt,right=10pt,top=8pt,bottom=10pt,
  title=\textbf{#1},
  fonttitle=\bfseries,
  coltitle=black,
  boxed title style={
    colback=titlegray,
    colframe=titlegray,
    arc=0pt,
    left=8pt,right=8pt,top=4pt,bottom=4pt
  }
}
\newenvironment{PromptBody}
{%
  \parindent=0pt
  \parskip=2pt
  \raggedright
  \sloppy
  \hyphenpenalty=10000
  \exhyphenpenalty=10000
}
{}

\title{MAGIC: Marginal-Guided Compression with Optimal Transport for Efficient Visual Document Retrieval}

\ificlrfinal
\author{
    Xu Yuan\textsuperscript{\rm 1},
    Hua Liu\textsuperscript{\rm 2},
    Wenqi Fan\textsuperscript{\rm 1}\thanks{Corresponding Author.},
    Qing Li\textsuperscript{\rm 1}\\
    \textsuperscript{\rm 1} The Hong Kong Polytechnic University, HK SAR\\
    \textsuperscript{\rm 2} Southern University of Science and Technology, China\\
    \texttt{xuyuan127@gmail.com},
    \texttt{liuh5@sustech.edu.cn}\\
    \texttt{wenqifan03@gmail.com},
    \texttt{qing-prof.li@polyu.edu.hk}
}
\else
\author{Anonymous Author(s)}
\fi

\begin{document}

\maketitle
\ificlrpreprint
  \lhead{Preprint}
\fi

\begin{abstract}
Recent visual document retrieval (VDR) systems such as ColPali use multi-vector page embeddings, in which patch-level vectors enable fine-grained evidence matching but incur substantial index storage and MaxSim scoring overhead.
Post-hoc merging offers a practical route to efficient VDR by reducing this cost without retraining the retriever, but its uniform reconstruction objectives are poorly aligned with the sparse, non-uniform patch usage induced by late-interaction retrieval.
Under aggressive compression, this misalignment can preserve rarely used patches while concentrating retrieval activity on too few retained representatives.
To address this misalignment, we propose \bfunderline{Ma}rginal-\bfunderline{G}u\bfunderline{i}ded \bfunderline{C}ompression with Optimal Transport (MAGIC), a training-free post-hoc compressor for efficient retrieval with frozen multi-vector embeddings.
MAGIC derives a MaxSim-induced compression surrogate and optimizes it through a two-marginal entropic optimal-transport formulation, where a retrieval-demand source marginal prioritizes high-use patches and a balanced target marginal regularizes retained-facet usage.
Across ViDoRe benchmarks, keep ratios, and retrieval backbones, MAGIC consistently outperforms strong post-hoc compressors, with particularly large gains in the aggressive-compression regime; component ablations verify the complementary effects of its two marginals. We release the code at: https://github.com/xandery-geek/MAGIC.
\end{abstract}
\section{Introduction}
\label{sec:intro}

Visual document retrieval (VDR) searches page images in which the layout and embedded visual elements carry evidence that plain-text retrieval may miss. 
It supports visual document QA~\citep{masry2022chartqa,tanaka2023slidevqa}, long-document analysis~\citep{ma2024mmlongbench,jiang2025hibench}, and retrieval-augmented generation~\citep{yu2025visrag,tanaka2025vdocrag}. 
Recent vision-language models (VLMs)~\citep{liu2023visual,bai2025qwen25vltechnicalreport,yuan2025instruction} make it possible to encode page images directly, reducing dependence on OCR-centric pipelines while preserving visual-layout evidence~\citep{gu2026unime,wei2025deepseek,wang2026render}.
In VDR, this shift is exemplified by ColPali~\citep{faysse2024colpali}, which instantiates multi-vector retrieval by representing each page with patch-level vectors and using late-interaction MaxSim to match each query vector to its most similar patch.
This fine-grained matching is effective but costly: each page stores hundreds of patch vectors, and online retrieval compares each query vector against them.

This cost has motivated efficient VDR methods that reduce the document-side vector count while preserving the fine-grained evidence that MaxSim uses. 
Learned compression methods train compact representations or modify the retrieval model, for example, through learned meta tokens~\citep{xiao2025metaembed} or modality-agnostic compression mechanisms~\citep{qin2026multi}. 
Post-hoc methods instead keep the VLM retriever frozen and compress the produced patch vectors, either by pruning low-value patches~\citep{yan2025docpruner}, merging patches into fewer representatives~\citep{ma2025towards}, or combining pruning and merging~\citep{yan2026sculpting}. 
The post-hoc regime is attractive for VDR because compression occurs after embedding extraction and can be applied to existing retrievers without retraining them.

Among post-hoc strategies, merging preserves retrieval evidence better than pruning at matched budgets~\citep{ma2025towards}. However, existing merging objectives remain retrieval-misaligned, and their retrieval quality degrades sharply when the retained-vector budget becomes tight.
These methods compress a page by grouping similar patches into retained representatives, which we refer to as facets, under a uniformly weighted reconstruction objective. Such query-agnostic reconstruction induces two capacity-allocation failures.
First, uniform patch weighting wastes capacity on patches rarely selected by query tokens under MaxSim.
Second, unconstrained facet usage can concentrate query-relevant patches on only a few retained facets, underusing the compressed vector budget.

\begin{wrapfigure}{r}{0.55\textwidth}
    \centering
    \includegraphics[page=1,width=\linewidth]{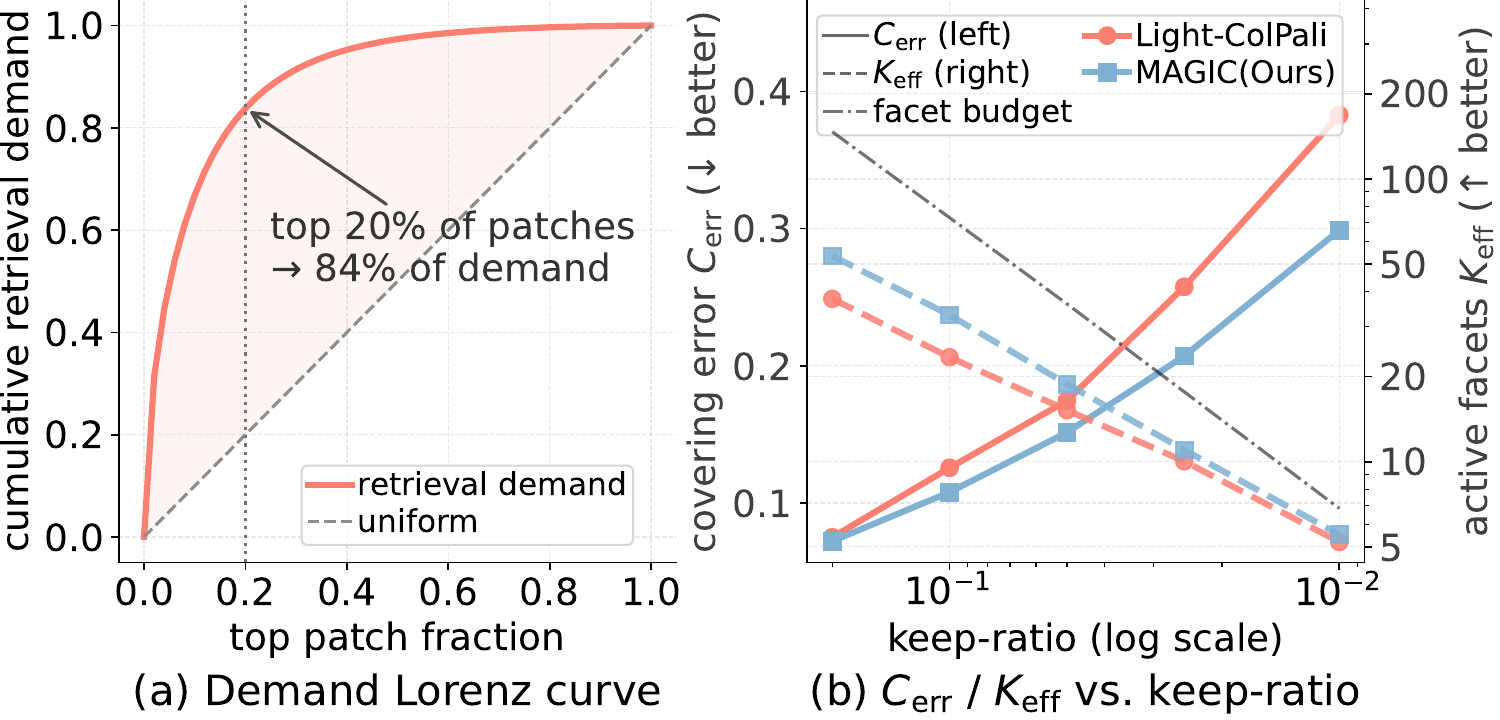}
    \caption{
    (a) Patch-level retrieval demand is concentrated: the top 20\% of patches account for about 84\% of query-token selections.
    (b) MAGIC reduces retrieval-demand-weighted covering error \(C_{\mathrm{err}}\) as the facet budget shrinks and maintains a higher retrieval-active facet count \(K_{\mathrm{eff}}\) than generic merging.}
    \label{fig:motivation}
\end{wrapfigure}
\noindent
We diagnose this misallocation by defining retrieval demand as how often each patch is selected by query tokens under MaxSim.
As illustrated in Figure~\ref{fig:motivation}(a), this demand is sharply concentrated: the top 20\% of patches account for approximately 84\% of all query-token selections. 
Using the same demand, Figure~\ref{fig:motivation}(b) shows that as the facet budget shrinks, a representative merging baseline incurs rapidly increasing retrieval-demand-weighted covering error \(C_{\mathrm{err}}\) while its retrieval-active facet count \(K_{\mathrm{eff}}\) remains far below the allocated budget.
Together, these diagnostics motivate a two-sided allocation objective: patch weights should follow retrieval demand, while retained facets should remain broadly retrieval-active.

Therefore, we propose \bfunderline{Ma}rginal-\bfunderline{G}u\bfunderline{i}ded \bfunderline{C}ompression with Optimal Transport (MAGIC), a training-free post-hoc compressor for efficient retrieval with frozen multi-vector embeddings.
To formalize the two-sided allocation, we first derive a MaxSim-induced compression surrogate showing that retrieval score degradation depends on demand-weighted patch coverage by the retained facets.
MAGIC instantiates this surrogate as a two-marginal optimal transport problem: the source marginal weights input patches by estimated retrieval demand, while the balanced target marginal spreads capacity across retained facets.
The resulting compressor runs offline during indexing, leaving online retrieval as standard MaxSim over the compressed facets, as shown in Figure~\ref{fig:framework}.

MAGIC improves the same diagnostic in Figure~\ref{fig:motivation}(b), reducing retrieval-demand-weighted covering error and increasing the retrieval-active facet count under tight facet budgets.
On the ViDoRe~\citep{faysse2024colpali} benchmark, these diagnostic gains translate to higher retrieval accuracy: MAGIC achieves the best average nDCG@5 at both moderate compression (\(r=0.1\)) and aggressive compression (\(r=0.01\)), with a larger margin when the retained-vector budget is most constrained.
Matched ablations further show that the retrieval-demand source marginal and the balanced target marginal provide complementary gains.

Our main contributions are highlighted as follows:
\begin{itemize}
    \item We diagnose retrieval misalignment in query-agnostic post-hoc merging, showing that uniform patch weighting overlooks concentrated retrieval demand and that unconstrained facet usage can collapse retrieval activity onto too few retained facets.
    \item We propose MAGIC, a training-free marginal-guided compressor that formulates post-hoc merging as a two-marginal optimal transport problem, combining a calibration-based retrieval-demand source marginal with a balanced target marginal.
    \item Extensive experiments across keep ratios, retrieval backbones, and ViDoRe benchmarks demonstrate consistent gains over strong post-hoc compressors, especially under aggressive compression, with ablations confirming the complementary effects of the two marginals.
\end{itemize}

\section{Related Work}
\label{sec:related}

\noindent \textbf{Visual Document Retrieval (VDR).} 
Visual document retrieval serves corpora whose semantics depend on page-level visual structure, supporting document QA and multimodal RAG~\citep{cho2024m3docrag,yuan2026mkg,jiang2026superglasses} over long or multi-page documents. 
Early pipelines often rely on OCR-derived text and text retrievers~\citep{karpukhin2020dense,wang2024improving}, which can miss visual-layout evidence.
VLM-based single-vector retrievers encode document images directly~\citep{ma2024unifying,zhang2025bridging,lin2025mm,gu2026unime}, reducing dependence on OCR but representing each page with a coarse global embedding.
ColPali~\citep{faysse2024colpali} extends the late-interaction multi-vector paradigm~\citep{khattab2020colbert,santhanam2022colbertv2} to VDR by representing each page with visual patch embeddings and using MaxSim-style scoring for fine-grained matching.
Subsequent work~\citep{gunther2025jina,nomicembedmultimodal2025} has further developed multi-vector retrievers for VDR.

\noindent \textbf{Efficient VDR.}
The storage and scoring cost of multi-vector VDR has motivated methods that
reduce the document-side vector count.
Learned efficient methods train compact token sets or lightweight compression
modules~\citep{xiao2025metaembed,qin2026multi}, but require additional training
or modifications to the retriever.
Post-hoc methods instead operate on frozen embeddings, including patch
pruning~\citep{yan2025docpruner,kankanampati2026voronoi}, vector
merging~\citep{ma2025towards,yan2026visual}, and prune-then-merge
pipelines~\citep{yan2026sculpting}.
This separation from retriever training makes post-hoc methods attractive for frozen VDR backbones; MAGIC follows the merging route but replaces query-agnostic reconstruction with a retrieval-aligned marginal formulation.

\noindent \textbf{Optimal Transport.}
Optimal transport provides a principled framework for assignment problems with prescribed source and target marginals~\citep{cuturi2013sinkhorn}.
It has been used for clustering~\citep{he2025dual,he2026prototype}, representation learning~\citep{caron2020unsupervised,yuan2023semantic}, domain adaptation~\citep{han2026vision}, and generative modeling~\citep{chiang2026cotem}.
Our work uses optimal transport as a compression mechanism for frozen
multi-vector VDR embeddings, where the prescribed marginals encode retrieval
demand and retained-vector capacity.
\section{Preliminaries}
\label{sec:prelim}

\subsection{Problem Setup}

We study post-hoc compression of frozen patch-level document embeddings.
For a page, let \(D=\{d_j\}_{j=1}^{N}\), \(d_j\in\mathbb{R}^{p}\), denote the patch embeddings produced by a fixed visual document retriever. 
A query is represented by token embeddings \(Q=\{q_i\}_{i=1}^{M}\). Late interaction~\citep{khattab2020colbert} scores the page by
\begin{equation}
    \operatorname{MaxSim}(Q,D)
    = \sum_{i=1}^{M}\max_{1\le j\le N} q_i^\top d_j .
    \label{eq:maxsim}
\end{equation}
The compression goal is to replace \(D\) with \(K\ll N\) retained facets \(F=\{f_k\}_{k=1}^{K}\), \(f_k\in\mathbb{R}^{p}\), while preserving the retrieval scores for future queries, i.e., \(\operatorname{MaxSim}(Q,F) \approx \operatorname{MaxSim}(Q,D)\).  
We set \(K=\min\{N,\max(1,\lceil rN\rceil)\}\) for keep-ratio \(r\), and construct one fixed compressed representation per document during offline indexing.

\subsection{Retrieval Misalignment in Generic Merging}
Generic merging compressors assign each patch embedding to one of \(K\) retained facets and construct each facet by aggregating its assigned patches.
Existing objectives~\citep{ma2025towards,yan2026sculpting} typically minimize a query-agnostic reconstruction loss whose uniform patch weights and unconstrained facet usage are misaligned with MaxSim behavior.

To diagnose this issue, we sample document pages from the ViDoRe benchmark~\citep{faysse2024colpali}. 
For each page, we generate five diverse queries with Qwen3-VL-8B~\citep{bai2025qwen3} conditioned on the page image, embed their query tokens with the same retriever, and record the patch selected by MaxSim for each token. 
Accumulating these selections gives a page-level retrieval-demand distribution. 
Using Light-ColPali~\citep{ma2025towards} as a representative merging baseline, we compute two demand-aware diagnostics: retrieval-demand-weighted covering error \(C_{\mathrm{err}}\), which measures facet coverage under this demand, and the effective number of retrieval-active facets \(K_{\mathrm{eff}}\), computed from the entropy of the per-facet retrieval-demand distribution. 
Formal definitions are given in Appendix~\ref{app:diagnostics}.
Figure~\ref{fig:motivation} shows that as the facet budget decreases, Light-ColPali exhibits increasing \(C_{\mathrm{err}}\), while \(K_{\mathrm{eff}}\) remains far below the allocated budget.
This diagnosis shows that generic merging misallocates the scarce facet budget on both sides of the compression map: source patches are weighted uniformly, while retained facet masses remain unconstrained.

\begin{figure*}[t]
    \centering
    \includegraphics[page=1,width=0.98\textwidth]{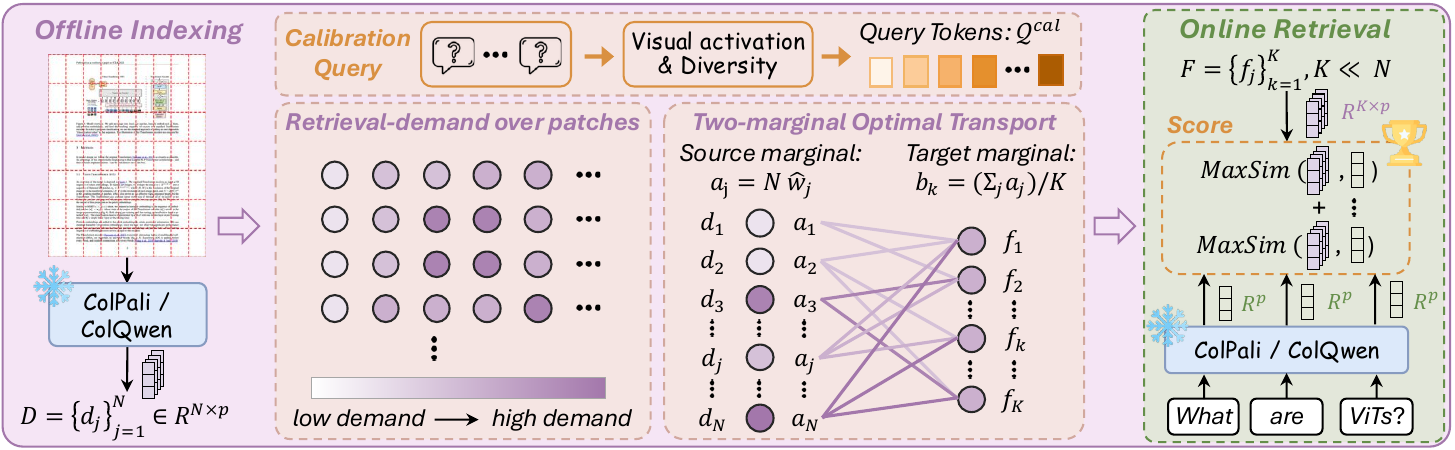}
    \caption{Overview of MAGIC. During offline indexing, frozen patch-level document embeddings are paired with calibration query tokens to estimate MaxSim-induced retrieval demand \(\widehat{w}_j\). MAGIC uses this demand as the source marginal and imposes a balanced target marginal in a two-marginal optimal transport problem, producing fixed compressed facets for online retrieval.}
    \label{fig:framework}
    \vspace{-1em}
\end{figure*}

\section{MAGIC}
\label{sec:method}
\label{sec:formulation}

Marginal-Guided Compression (MAGIC) addresses this misallocation by aligning both sides of the compression map with retrieval use: patch-side mass follows retrieval demand, while retained-facet capacity is balanced across facets. 
We implement this principle as a two-marginal optimal transport problem, as shown in Figure~\ref{fig:framework}.

\subsection{MaxSim-Induced Compression Surrogate}
This section first formalizes the patch-side objective. Motivated by the MaxSim scoring scheme, we derive a compression surrogate that upper bounds the MaxSim degradation caused by compression and identifies retrieval demand as the appropriate source weighting.

\paragraph{MaxSim-Induced Source Marginal.}
We now define the population counterpart of the retrieval-demand distribution
used in the diagnosis. 
For a query token \(q\), MaxSim selects the patch \(j^*(q)=\operatorname*{arg\,max}_j q^\top d_j\). 
Thus, the retrieval demand of a document patch is the probability that future query tokens select it. Let \(\mathcal{P}_Q\) denote the population distribution of query-token embeddings for the retrieval workload, and define
\begin{equation}
    w_j=\Pr_{q\sim\mathcal{P}_Q}[j^*(q)=j],
    \qquad
    \sum_{j=1}^{N}w_j=1 .
    \label{eq:maxsim-source-marginal}
\end{equation}
The vector \(w\) is the MaxSim-induced selection distribution over document patches.

\paragraph{Surrogate Objective.}
\label{subsec:compression-bound}

With this selection distribution, we connect compression error to retrieval score degradation. 
A multi-token MaxSim score is a sum of token-level maxima, so the full-query positive degradation is bounded by summing token-level positive decreases.
Given a query token \(q\), let \(h_D(q)=\max_j q^\top d_j\) and \(h_F(q)=\max_k q^\top f_k\) be its token-level MaxSim scores before and after compression.
The positive single-token MaxSim degradation is measured by \(\Delta_+(q;D,F)=[h_D(q)-h_F(q)]_+\), where \([x]_+=\max\{x,0\}\) keeps only score decreases caused by compression.

\noindent\emph{Proposition 1.}
Let \(D=\{d_j\}_{j=1}^{N}\) and \(F=\{f_k\}_{k=1}^{K}\) contain unit-normalized vectors, and let \(\pi(j)\) map each document patch to a compressed facet. 
For query tokens \(q\sim\mathcal{P}_Q\) with \(\lVert q\rVert_2\le1\), and for \(w\) defined in Eq.~\ref{eq:maxsim-source-marginal}, we have
\begin{equation}
\begin{aligned}
    \mathbb{E}_q[\Delta_+(q;D,F)]
    \le
    \sum_{j=1}^{N} w_j \lVert d_j-f_{\pi(j)}\rVert_2
    \le
    \sqrt{2\sum_{j=1}^{N}w_j
    \left(1-d_j^\top f_{\pi(j)}\right)} .
\end{aligned}
\label{eq:weighted-bound}
\end{equation}

\noindent\emph{Proof sketch.}
For a fixed query token, abbreviate \(j^*=j^*(q)\), so \(d_{j^*}\) is the patch selected by MaxSim and \(f_{\pi(j^*)}\) is the facet assigned to this selected patch. Because \(h_F(q)\) maximizes over all facets, any positive score decrease is no larger than the loss from replacing \(d_{j^*}\) with \(f_{\pi(j^*)}\):
\begin{equation}
\begin{aligned}
    \Delta_+(q;D,F)
    \le
    \left[q^\top d_{j^*}-q^\top f_{\pi(j^*)}\right]_+
    \le
    \left|q^\top(d_{j^*}-f_{\pi(j^*)})\right|
    \le
    \lVert d_{j^*}-f_{\pi(j^*)}\rVert_2 .
\end{aligned}
\label{eq:proof-sketch}
\end{equation}
The last step uses Cauchy--Schwarz inequality and \(\lVert q\rVert_2\le1\). Taking expectation and grouping query tokens by the event \(j^*(q)=j\) yields the first inequality in Eq.~\ref{eq:weighted-bound}. 
The second follows from the weighted root-mean-square inequality and, for normalized vectors, \(\lVert d_j-f_{\pi(j)}\rVert_2^2=2(1-d_j^\top f_{\pi(j)})\). 
A complete proof, including the multi-token extension, is provided in Appendix~\ref{app:compression-bound-proof}.

Since the square root in Eq.~\ref{eq:weighted-bound} is monotone, reducing the inner demand-weighted reconstruction cost tightens the upper bound.
This motivates the retrieval-demand-weighted compression surrogate
\begin{equation}
    \mathcal{L}_{\mathrm{RD}}(F,\pi)
    =
    \sum_{j=1}^{N}w_j
    \left(1-d_j^\top f_{\pi(j)}\right).
    \label{eq:rd-surrogate}
\end{equation}
Patches with little retrieval demand have \(w_j\approx0\), so their reconstruction errors contribute little to the surrogate. High-demand patches receive larger weights and must be preserved more accurately.
Thus, the uniform objective in generic merging is a loose proxy for MaxSim preservation because it penalizes errors on low- and high-demand patches equally.


\subsection{Marginal-Guided Compressor}

We now convert the surrogate in Eq.~\ref{eq:rd-surrogate} into a compressor.
First, it is rewritten as a source-weighted patch-to-facet assignment, where patch mass is given by retrieval demand.
Second, a target marginal is introduced to balance retained-facet usage, yielding a two-marginal OT formulation.
Writing the hard patch-to-facet map \(\pi\) with assignment indicators \(Z_{kj}=\mathbf{1}[k=\pi(j)]\), where \(\sum_{k=1}^{K}Z_{kj}=1\), and using \(c(f_k,d_j)=1-d_j^\top f_k\), we have an assignment-form rewriting
\begin{equation}
    N\mathcal{L}_{\mathrm{RD}}(F,\pi)
    =
    \sum_k\sum_j
    Nw_j Z_{kj}\,c(f_k,d_j).
    \label{eq:weighted-assignment-objective}
\end{equation}
Here, the factor \(N\) only rescales the objective and sets the total assignment mass to \(N\), without changing the minimizer.
Thus, minimizing the right-hand side of Eq.~\ref{eq:weighted-assignment-objective} is equivalent to minimizing the MaxSim-induced surrogate in Eq.~\ref{eq:rd-surrogate}.
Let \(a_j=Nw_j\) and absorb this source mass into the assignment by setting \(T_{kj}=a_jZ_{kj}\). Then \(\sum_kT_{kj}=a_j\) and \(T_{kj}\in\{0,a_j\}\). Here \(F\) contains the retained facets to be learned, whereas \(T\) assigns source mass to facets for a given \(F\). 
Defining \(\mathcal{H}(a)=\{T:\sum_kT_{kj}=a_j,\ T_{kj}\in\{0,a_j\}\}\), the equivalent hard source-constrained assignment is
\begin{equation}
    \min_F\;\min_{T\in\mathcal{H}(a)}
    \sum_k\sum_jT_{kj}c(f_k,d_j).
    \label{eq:hard-transport-objective}
\end{equation}
The discrete constraint requires choosing one of \(K\) facets for each of the \(N\) patches while simultaneously learning the facet vectors, making exact optimization combinatorial. MAGIC therefore relaxes \(T_{kj}\in\{0,a_j\}\) to \(T_{kj}\ge0\) and adds entropy, yielding a tractable soft-assignment problem.

\paragraph{Two-Marginal Optimal Transport.}
In the relaxed problem, \(T_{kj}\) is the mass transported from patch \(d_j\) to facet \(f_k\). The vector \(a\) has entries \(a_j=Nw_j\), assigning patch mass according to retrieval demand.
MAGIC further prescribes a balanced target marginal \(b_k=(\sum_j a_j)/K\) over retained facets. 
Let \(\mathcal{U}(a,b)=\{T\ge0:\sum_kT_{kj}=a_j,\ \sum_jT_{kj}=b_k\}\). For fixed facets, the inner problem over \(T\) is entropic OT. Learning \(F\) gives the compressor objective
\begin{equation}
\begin{aligned}
    \min_F\;\min_{T\in\mathcal{U}(a,b)}\quad
    \sum_{k=1}^{K}\sum_{j=1}^{N} T_{kj}\,c(f_k,d_j) 
    +\varepsilon\sum_{k=1}^{K}\sum_{j=1}^{N}T_{kj}(\log T_{kj}-1),
\end{aligned}
\label{eq:magic-objective}
\end{equation}
which extends the hard source-constrained assignment in Eq.~\ref{eq:hard-transport-objective} into a two-marginal entropic OT: \(a\) preserves source-side retrieval demand, \(b\) balances retained-facet capacity, and the entropy term yields a stable soft transport plan.

The population retrieval-demand distribution \(w\) in Eq.~\ref{eq:maxsim-source-marginal} is unknown at indexing time, so MAGIC estimates it from a held-out calibration query-token pool \(\mathcal{Q}^{\mathrm{cal}}\), constructed from training queries disjoint from evaluation queries. 
For each document, MAGIC computes the smoothed MaxSim selection estimate
\begin{equation}
    \widehat{w}_j(\mathcal{Q}^{\mathrm{cal}},\tau)
    =
    \frac{1}{|\mathcal{Q}^{\mathrm{cal}}|}
    \sum_{q\in\mathcal{Q}^{\mathrm{cal}}}
    \frac{\exp(q^\top d_j/\tau)}
    {\sum_{\ell=1}^{N}\exp(q^\top d_{\ell}/\tau)}.
    \label{eq:selection-estimator}
\end{equation}
The softmax form yields a positive estimate \(\widehat{w}\), and in practice, MAGIC sets \(a_j=N\widehat{w}_j\), matching the total source mass in Eq.~\ref{eq:magic-objective}.
The temperature \(\tau\) smooths the estimate; as \(\tau\to0\), \(\widehat{w}_j\) approaches the empirical frequency with which patch \(j\) is selected by MaxSim.

For the target side, the preliminary diagnostic motivates preventing retained facets from absorbing capacity arbitrarily. This is why we use the balanced target marginal defined above.
The constraint prevents the soft assignment from concentrating transported mass on only a few retained facets, ensuring efficient use of the compression budget. This balance regularizes the soft transport plan without enforcing equal final cluster sizes.

\paragraph{Calibration Query Prior Construction.}
The calibration query-token pool \(\mathcal{Q}^{\mathrm{cal}}\) is constructed once offline from held-out training queries. 
To bias this pool toward visually grounded query tokens, we first build a corpus visual dictionary by uniformly sampling training document pages, encoding them with the frozen retriever into patch embeddings, and selecting \(P\) diverse representatives from the collected patches by farthest-first selection~\citep{arthur2007k}. 
Let the resulting dictionary be \(\mathcal{V}_{\mathrm{corp}}\). For a normalized query token \(\widehat{q}\), define its visual activation as \(\operatorname{vis}(q)= \max\{0,\max_{v\in\mathcal{V}_{\mathrm{corp}}}\widehat{q}^{\top}v\}\). 
Starting from the token with the largest visual activation, we select \(L_{\mathrm{cal}}\) normalized calibration tokens from the held-out query-token pool by score-biased farthest-first selection:
\begin{equation}
    q^*
    =
    \operatorname*{arg\,max}_{q_j\notin S}
    \operatorname{vis}(q_j)
    \min_{s\in S}\left(1-\widehat{q}_j^{\top}\widehat{s}\right),
    \label{eq:score-biased-fps}
\end{equation}
where \(S\) is the current selected set. This criterion keeps visually activated tokens while preserving diversity in query-token space. 
Notably, the resulting \(\mathcal{Q}^{\mathrm{cal}}\) is used only to estimate \(\widehat{w}\) offline.

\paragraph{Optimization Algorithm.}

To optimize Eq.~\ref{eq:magic-objective} with fixed marginals, MAGIC alternates between the inner transport plan \(T\) and the learned facets \(F\).
We initialize \(F\) by deterministic farthest-first seeding over the normalized patch vectors. 
With facets fixed, define \(M_{kj}=f_k^\top d_j/\varepsilon\), and let \(u\in\mathbb{R}^{K}\) and \(v\in\mathbb{R}^{N}\) be the log-domain Sinkhorn potentials for the target and source marginals, respectively. 
The constant term in \(c(f_k,d_j)=1-f_k^\top d_j\) is absorbed into these potentials. Log-domain Sinkhorn~\citep{cuturi2013sinkhorn} updates compute the transport plan:
\begin{equation}
\begin{aligned}
    u_k \leftarrow \log b_k-\operatorname{logsumexp}_{j}(M_{kj}+v_j), &\quad
    v_j \leftarrow \log a_j-\operatorname{logsumexp}_{k}(M_{kj}+u_k),\\
    T_{kj} &= \exp(M_{kj}+u_k+v_j).
\end{aligned}
\label{eq:sinkhorn}
\end{equation}
With \(T\) fixed, we row-normalize it to \(W_{kj}=T_{kj}/\sum_{j'}T_{kj'}\) and update each facet by a damped spherical barycenter step:
\begin{equation}
    f_k \leftarrow
    \operatorname{norm}\!\left(
    (1-\eta)f_k+\eta\sum_{j=1}^{N}W_{kj}d_j
    \right).
    \label{eq:barycenter-update}
\end{equation}
All patch embeddings are normalized before compression, and facet vectors are normalized after initialization and after each barycenter update. 
After the final transport iteration, MAGIC applies a spherical-Lloyd readout~\citep{dhillon2001concept}, which assigns each patch to its most similar learned facet and takes the retrieval-demand-weighted normalized mean:
\begin{equation}
    k(j)=\operatorname*{arg\,max}_{k} f_k^\top d_j,\quad
    f_k^{\mathrm{out}}=\operatorname{norm}\!\left(\sum_{k(j)=k} a_j d_j\right).
    \label{eq:lloyd-readout}
\end{equation}
If no patch is assigned to a facet, MAGIC keeps its pre-readout transport barycenter. Finally, \(F^{\mathrm{out}}=\{f_k^{\mathrm{out}}\}_{k=1}^{K}\) is stored as the
compressed set for subsequent MaxSim retrieval. The complete algorithm procedure is provided in Appendix~\ref{app:magic-algorithm}.

\begin{table*}[t]
\caption{Comparison with post-hoc baselines on ViDoRe v1 using ColQwen2.5.
The metric is nDCG@5 (\%) / Recall@5 (\%). The upper block reports moderate compression
(\(r=0.1\)); the lower block reports aggressive compression (\(r=0.01\)).}
\label{tab:posthoc-main}
\centering
\setlength{\tabcolsep}{3pt}
\resizebox{\textwidth}{!}{
\begin{tabular}{lccccccc}
\toprule
\textbf{Method} & \textbf{ArxivQ} & \textbf{DocQ} & \textbf{InfoQ} & \textbf{TabF} & \textbf{TATQ} & \textbf{Shift} & \textbf{Avg.} \\
\midrule
\multicolumn{8}{c}{\textbf{Keep-ratio \(r=0.1\)}} \\
1D-Pooling       & 80.32 / 85.20 & 52.80 / 58.24 & 86.96 / 91.34 & 85.20 / 89.64 & 72.29 / 83.78 & 71.43 / 85.00 & 74.83 / 82.20 \\
2D-Pooling       & 82.28 / 86.80 & 53.85 / 60.12 & 86.80 / 91.09 & 86.20 / 91.79 & 67.16 / 79.04 & 74.47 / 88.00 & 75.13 / 82.81 \\
K-Means          & 86.42 / 90.40 & 58.66 / 64.81 & 89.31 / 93.73 & 88.90 / 93.93 & 77.40 / 87.18 & 80.07 / 90.00 & 80.13 / 86.67 \\
DocPruner        & 74.68 / 81.00 & 47.40 / 55.61 & 77.32 / 84.05 & 85.47 / 92.86 & 60.95 / 72.27 & 54.32 / 67.00 & 66.69 / 75.46 \\
Voronoi Pruning  & 82.88 / 87.00 & 53.15 / 60.49 & 78.08 / 85.36 & 88.27 / 92.86 & 66.04 / 76.91 & 56.03 / 73.00 & 70.74 / 79.27 \\
Light-ColPali    & 86.58 / 90.80 & 59.15 / 66.25 & 90.66 / \textbf{94.17} & \textbf{89.45} / 95.00  & 78.94 / 88.70 & 83.17 / 93.00 & 81.32 / 87.99 \\
ColChunk         & 85.19 / 89.40 & 58.27 / 66.29 & 88.12 / 92.75 & 88.79 / 94.29 & 76.82 / 87.21 & 77.33 / 87.00 & 79.09 / 86.16 \\
Prune-then-Merge & 84.49 / 89.20 & 59.26 / 67.01 & 87.61 / 92.31 & 88.97 / \textbf{95.36} & 77.04 / 86.36 & 76.50 / 87.00 & 78.98 / 86.21 \\
\midrule
\rowcolor{gray!20} \textbf{MAGIC}   & \textbf{86.91 / 91.00} & \textbf{60.49 / 67.55} & \textbf{90.70} / 93.97 & 89.39 / 94.64 & \textbf{79.24 / 89.16} & \textbf{86.01 / 94.00} & \textbf{82.12 / 88.39} \\
\midrule
\multicolumn{8}{c}{\textbf{Keep-ratio \(r=0.01\)}} \\
1D-Pooling       & 66.63 / 74.00 & 37.16 / 44.42 & 79.20 / 85.53 & 75.60 / 83.21 & 50.93 / 62.94 & 55.67 / 73.00 & 60.87 / 70.52 \\
2D-Pooling       & 69.01 / 76.00 & 35.58 / 44.46 & 78.92 / 86.54 & 74.73 / 81.79 & 48.20 / 60.24 & 56.50 / 74.00 & 60.49 / 70.51 \\
K-Means          & 74.81 / 81.20 & 47.49 / 55.18 & 81.62 / 87.15 & 79.42 / 86.07 & 59.75 / 71.90 & 60.09 / 74.00 & 67.20 / 75.92 \\
DocPruner        & 29.98 / 37.20 & 21.46 / 27.15 & 45.65 / 53.85 & 46.74 / 54.64 & 28.90 / 37.21 & 17.96 / 22.00 & 31.78 / 38.68 \\
Voronoi Pruning  & 43.74 / 52.40 & 15.34 / 19.29 & 23.37 / 30.47 & 47.48 / 57.50 & 18.82 / 23.63 & 13.14 / 16.00 & 26.98 / 33.21 \\
Light-ColPali    & 77.32 / 83.40 & 45.51 / 52.36 & 80.84 / 87.45 & 81.73 / 88.93 & 60.51 / 72.63 & 64.17 / 78.00 & 68.34 / 77.13 \\
ColChunk         & 79.16 / \textbf{85.20} & 45.66 / 54.69 & 82.69 / 88.56 & 82.95 / 88.93 & 61.98 / 72.87 & 68.11 / 79.00 & 70.09 / 78.21 \\
Prune-then-Merge & 74.54 / 80.00 & 46.05 / 54.75 & 78.85 / 85.42 & 82.98 / 88.57 & 62.44 / 73.06 & 61.95 / 77.00 & 67.80 / 76.47 \\
\midrule
\rowcolor{gray!20} \textbf{MAGIC}   & \textbf{79.54} / 84.60 & \textbf{48.61 / 55.48} & \textbf{83.15 / 89.17} & \textbf{82.99 / 89.02} & \textbf{62.80 / 75.73} & \textbf{69.17 / 86.00} & \textbf{71.04 / 80.00} \\
\bottomrule
\end{tabular}
}
\vspace{-1em}
\end{table*}

\section{Experiments}
\label{sec:exp}

\subsection{Experimental Setup}
\label{subsec:exp-setup}

\noindent \textbf{Baselines and Benchmarks.}
We compare MAGIC with post-hoc compressors operating on frozen document embeddings, including 1D-Pooling, 2D-Pooling, K-Means, DocPruner~\citep{yan2025docpruner}, Voronoi Pruning~\citep{kankanampati2026voronoi}, Light-ColPali~\citep{ma2025towards}, ColChunk~\citep{yan2026visual}, and Prune-then-Merge~\citep{yan2026sculpting}. 
Unless otherwise specified, all post-hoc methods use ColQwen2.5 as the frozen retriever, isolating the effect of the compression strategy from embedding extraction.
We use ViDoRe v1~\citep{faysse2024colpali} as the primary benchmark and ViDoRe v2~\citep{mace2025vidore} for multilingual transfer.
We report nDCG@5 and Recall@5 as retrieval metrics.
Additional results on the reasoning-intensive ViDoSeek benchmark~\citep{wang2025vidorag} are provided in Appendix~\ref{app:generalization}.

\noindent \textbf{Implementation Details.}
Unless otherwise specified, all experiments use ColQwen2.5 as the frozen retriever.
For each page with \(N\) patch embeddings, the keep-ratio \(r\) defines a nominal retained-vector budget \(K=\min\{N,\max(1,\lceil rN\rceil)\}\).
Different compressors may realize this budget slightly differently, so post-hoc baselines are compared under approximately matched retained-vector counts.
All patch vectors are L2-normalized before compression, and online retrieval uses the standard MaxSim score over the retained vectors.
Across all main experiments, MAGIC uses \(L_{\mathrm{cal}}=1000\), \(\tau=0.05\), \(\varepsilon=0.05\), \(T_{\mathrm{out}}=5\), and \(T_{\mathrm{sk}}=5\).
Additional hyperparameters, preprocessing, and baseline-specific details are provided in Appendix~\ref{sec:app-experimental-details}.

\subsection{Main Results}
\label{subsec:main-results}

\noindent \textbf{Comparison with Post-hoc Baselines.}
We first compare MAGIC with post-hoc baselines under matched keep ratios on ViDoRe v1 with ColQwen2.5. 
This comparison evaluates compressors that take the same frozen page embeddings as input and use approximately matched retained-vector budgets. 
Table~\ref{tab:posthoc-main} reports nDCG@5 and Recall@5 at two representative budgets: \(r=0.1\), which measures moderate compression, and \(r=0.01\), which stresses the aggressive-compression regime where capacity allocation is most constrained. 
MAGIC achieves the best average performance at both budgets, reaching \(82.12/88.39\) at \(r=0.1\) compared with \(81.32/87.99\) for the strongest baseline, Light-ColPali. 
The advantage grows under aggressive compression: at \(r=0.01\), MAGIC obtains \(71.04/80.00\), exceeding ColChunk by \(0.95/1.79\) and Light-ColPali by \(2.70/2.87\). 
This fixed-ratio protocol isolates the compression strategy by using the same frozen retriever and approximately matched retained-vector budgets across post-hoc methods.

\begin{figure*}[t]
    \centering
    \begin{minipage}[t]{0.245\textwidth}
        \centering
        \includegraphics[width=\linewidth]{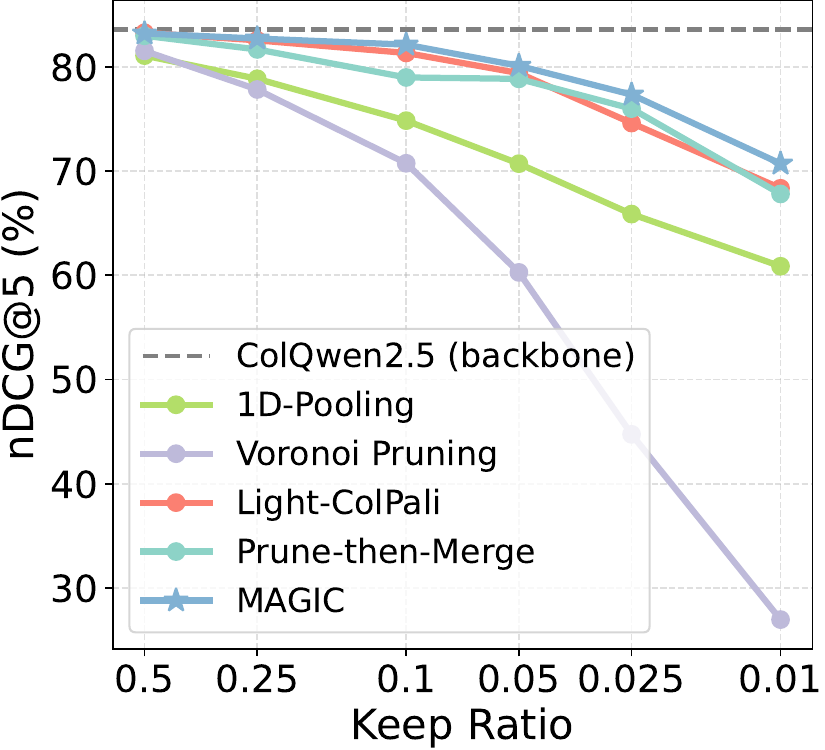}\\
        \scriptsize (a) ViDoRe v1 - ColQwen2.5
    \end{minipage}\hfill
    \begin{minipage}[t]{0.245\textwidth}
        \centering
        \includegraphics[width=\linewidth]{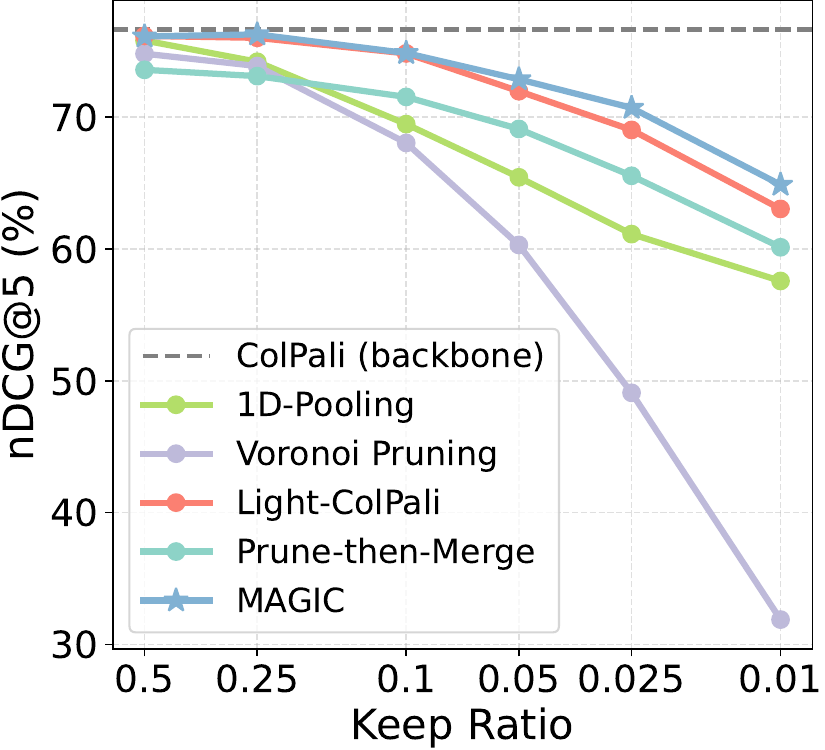}\\
        \scriptsize (b) ViDoRe v1 - ColPali
    \end{minipage}\hfill
    \begin{minipage}[t]{0.245\textwidth}
        \centering
        \includegraphics[width=\linewidth]{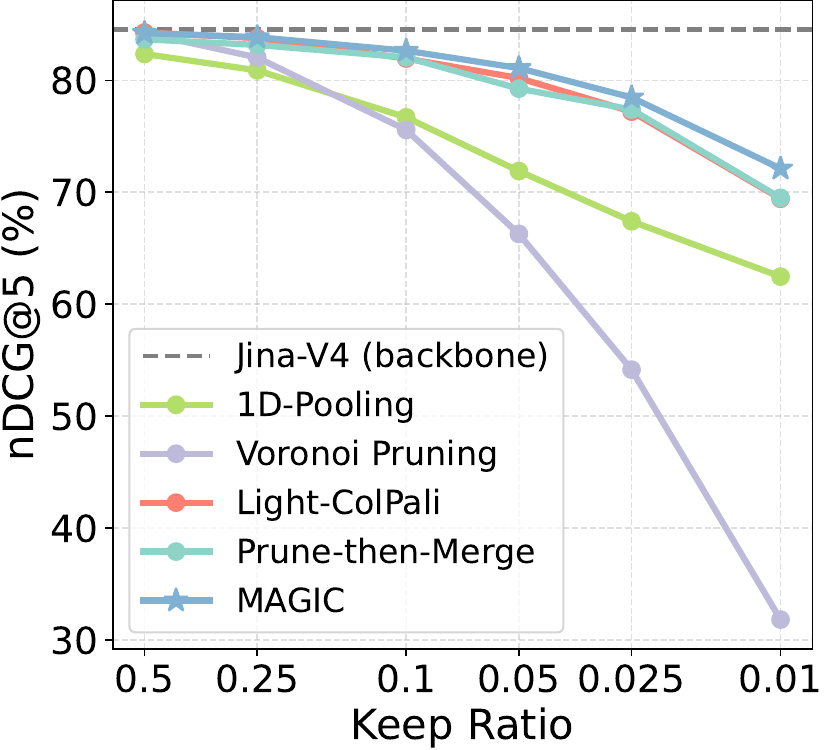}\\
        \scriptsize (c) ViDoRe v1 - Jina-v4
    \end{minipage}\hfill
    \begin{minipage}[t]{0.245\textwidth}
        \centering
        \includegraphics[width=\linewidth]{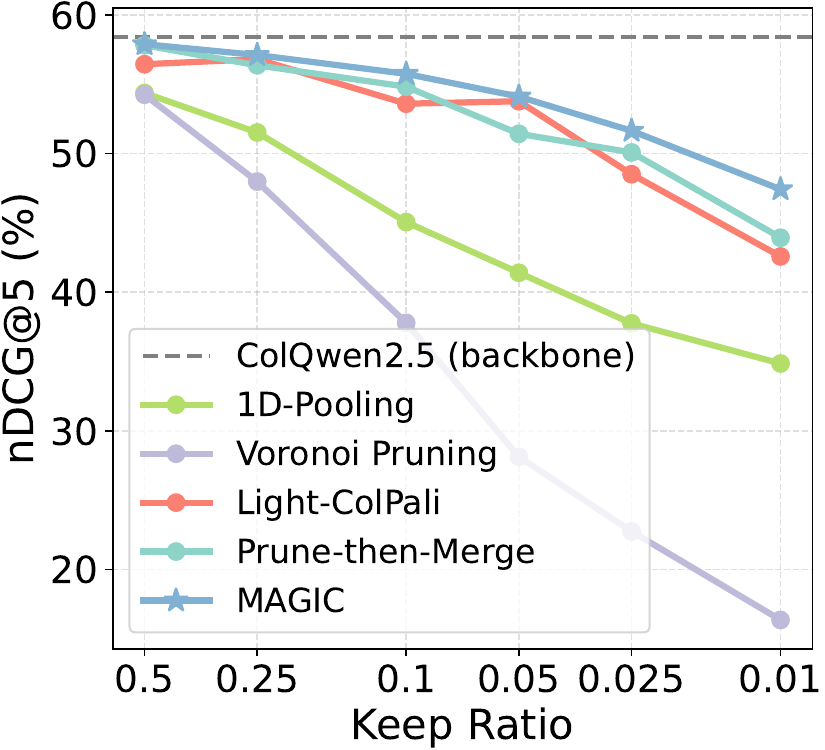}\\
        \scriptsize (d) ViDoRe v2 - ColQwen2.5
    \end{minipage}
    \caption{Budget-sweep retrieval performance (nDCG@5) across main evaluation settings.}
    \label{fig:dense-budget-curves}
    \vspace{-1em}
\end{figure*}

To examine performance beyond these two operating points, Figure~\ref{fig:dense-budget-curves}(a)
reports budget-sweep curves on ViDoRe v1 with ColQwen2.5 over a broad range of keep ratios.
The curve complements Table~\ref{tab:posthoc-main} by showing that MAGIC's gains are not tied to isolated budgets:
it matches or exceeds strong generic merging baselines at moderate budgets and separates more clearly as the retained-vector budget becomes scarce.
In Appendix~\ref{app:learned-baselines}, we further compare MAGIC with learned efficient baselines under a fixed budget.

\noindent \textbf{Generalization.}
In Figures~\ref{fig:dense-budget-curves}(b)--(d), we test whether the budget-sweep trend transfers across retrieval backbones and benchmarks.
On ViDoRe v1, the ColPali and Jina-v4~\citep{gunther2025jina} curves test backbone transfer beyond the default ColQwen2.5 retriever; on ViDoRe v2, the ColQwen2.5 curve tests multilingual benchmark transfer under the same retriever.
The consistent gains across these settings show that retrieval-aligned capacity allocation remains effective across different frozen retrievers and evaluation data.

\subsection{Ablation Study}
\label{subsec:ablation}

\begin{table}[t]
\caption{Marginal decomposition on ViDoRe v1. The metric is nDCG@5 (\%).
\ding{51} and \ding{55} indicate whether each component is used; disabling
Lloyd uses the soft barycenter readout.}
\label{tab:marginal-decomposition}
\centering
\resizebox{0.75\columnwidth}{!}{
\begin{tabular}{ccccccccc}
\toprule
\textbf{Source} & \textbf{Target} & \textbf{Lloyd} & \textbf{0.5} & \textbf{0.25} & \textbf{0.1} & \textbf{0.05} & \textbf{0.025} & \textbf{0.01} \\
\midrule
\ding{51} & \ding{51} & \ding{51} & \textbf{83.20} & \textbf{82.71} & \textbf{82.12} & \textbf{80.09} & \textbf{77.33} & \textbf{70.71} \\
\ding{55} & \ding{51} & \ding{51} & 82.41 & 81.67 & 79.11 & 76.70 & 71.60 & 61.88 \\
\ding{51} & \ding{55} & \ding{51} & 82.72 & 82.52 & 81.19 & 79.04 & 76.83 & 70.02 \\
\ding{51} & \ding{51} & \ding{55} & 82.62 & 82.36 & 81.71 & 79.99 & 77.01 & 70.19 \\
\bottomrule
\end{tabular}
}
\vspace{-1em}
\end{table}

\noindent \textbf{Marginal Decomposition.}
We isolate the contribution of each design choice in Table~\ref{tab:marginal-decomposition}, which decomposes the two marginal choices and the spherical-Lloyd readout. 
The full model achieves the highest average nDCG@5 across all keep ratios. 
The source marginal is most critical under tight budgets: replacing it with a uniform source drops performance from 82.12 to 79.11 at \(r=0.1\), and from 70.71 to 61.88 at \(r=0.01\). 
Removing the balanced target marginal or the Lloyd readout causes smaller but consistent declines, confirming that these components provide complementary gains.

\begin{figure}[t]
    \centering
    \begin{minipage}[t]{0.49\textwidth}
        \centering
        \begin{minipage}[t]{0.48\linewidth}
            \centering
            \includegraphics[width=\linewidth]{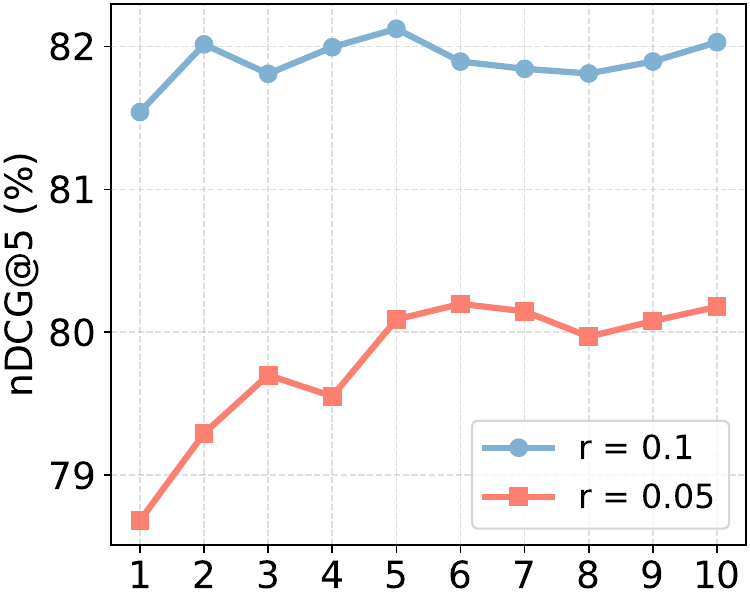}\\
            \small (a) Outer iterations \(T_{\mathrm{out}}\)
        \end{minipage}\hfill
        \begin{minipage}[t]{0.48\linewidth}
            \centering
            \includegraphics[width=\linewidth]{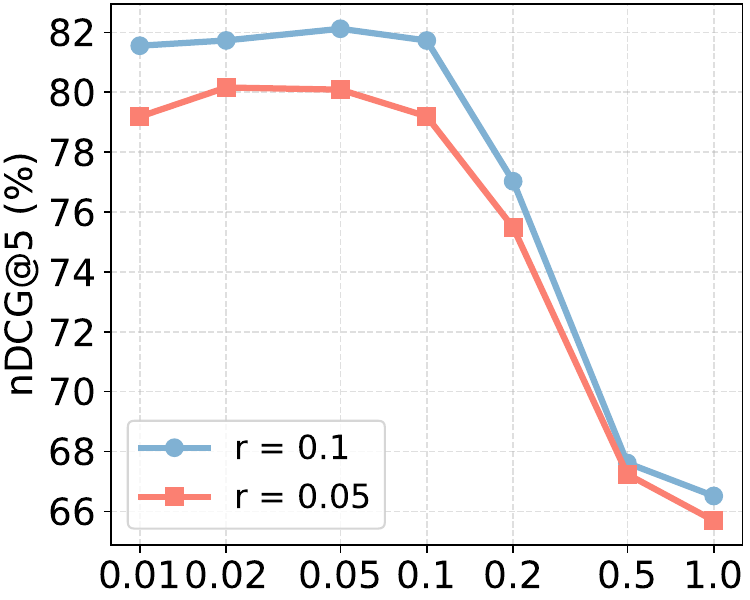}\\
            \small (b) Temperature \(\varepsilon\)
        \end{minipage}
        \caption{Sensitivity to optimization choices.}
        \label{fig:optimization-choices}
    \end{minipage}\hfill
    \begin{minipage}[t]{0.49\textwidth}
        \centering
        \begin{minipage}[t]{0.44\linewidth}
            \centering
            \includegraphics[width=\linewidth]{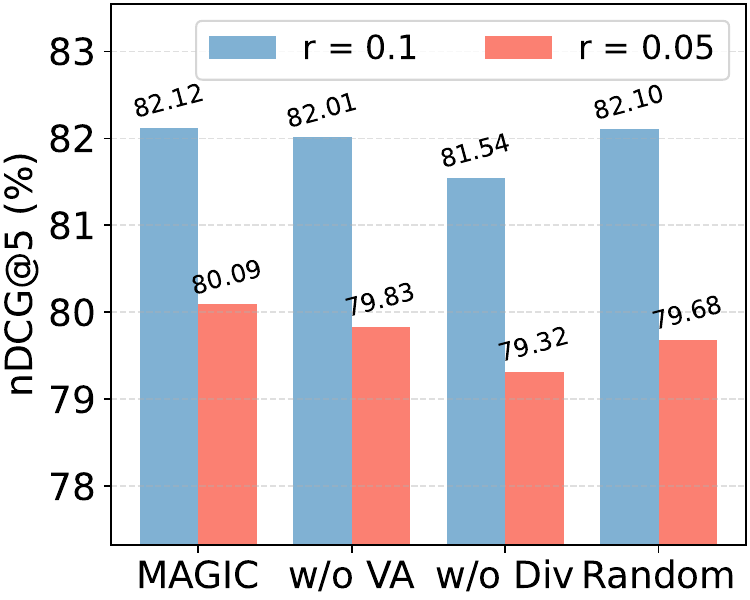}\\
            \small (a) Selection strategy
        \end{minipage}\hfill
        \begin{minipage}[t]{0.52\linewidth}
            \centering
            \includegraphics[width=\linewidth]{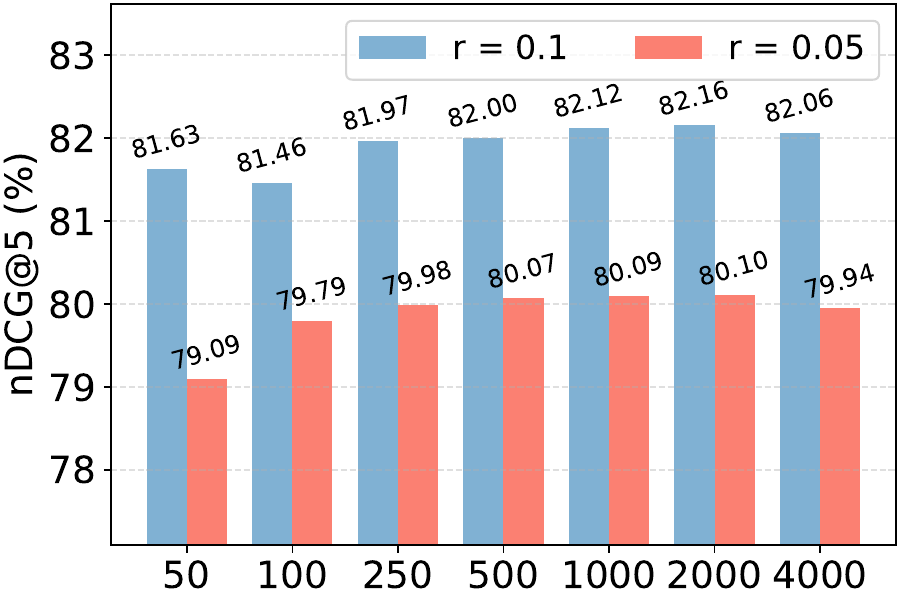}\\
            \small (b) Calibration token count
        \end{minipage}
        \caption{Calibration query-token pool ablations.}
        \label{fig:calibration-query-prior}
    \end{minipage}
    \vspace{-1em}
\end{figure}

\noindent \textbf{Optimization Choices.}
MAGIC remains stable without delicate optimization tuning, as shown in Figure~\ref{fig:optimization-choices}.
Performance improves after the first few outer iterations and then saturates. For the entropic temperature \(\varepsilon\), moderate values perform similarly, whereas overly large values make the transport plan too diffuse and reduce retrieval accuracy. More hyperparameter ablations are provided in Appendix~\ref{app:additional-hyperparameters}.

\noindent \textbf{Calibration Query-Token Pool.}
Figure~\ref{fig:calibration-query-prior} ablates how the calibration query-token pool is constructed for estimating the source marginal.
We compare MAGIC with three variants: \textit{w/o VA} removes visual activation and selects diverse tokens only, \textit{w/o Div} removes diversity and keeps the highest visual-activation tokens, and Random samples tokens uniformly.
MAGIC performs best, showing that visual grounding and token diversity are complementary. 
Varying the calibration size shows rapid saturation, with little change between 1000 and 2000 tokens. Thus, MAGIC obtains a reliable retrieval-demand estimate without requiring many calibration tokens.

\subsection{Discussion}
\label{subsec:discussion}

\noindent
\begin{minipage}[t]{0.4\textwidth}
    \vspace{0pt}
    \textbf{Interpretability.}
    Figure~\ref{fig:interpretability} illustrates whether MAGIC preserves query-grounded evidence after compression.
    For the uncompressed index, MaxSim directly identifies the original image patches activated by query tokens.
    For MAGIC, MaxSim selects compressed facets, which are back-projected to source patches using the learned assignment information.
    The highlighted regions remain concentrated around the answer field, suggesting that the compressed facets retain the document evidence used by late-interaction retrieval.
\end{minipage}\hfill
\begin{minipage}[t]{0.58\textwidth}
    \vspace{0pt}
    \centering
    \includegraphics[width=\linewidth]{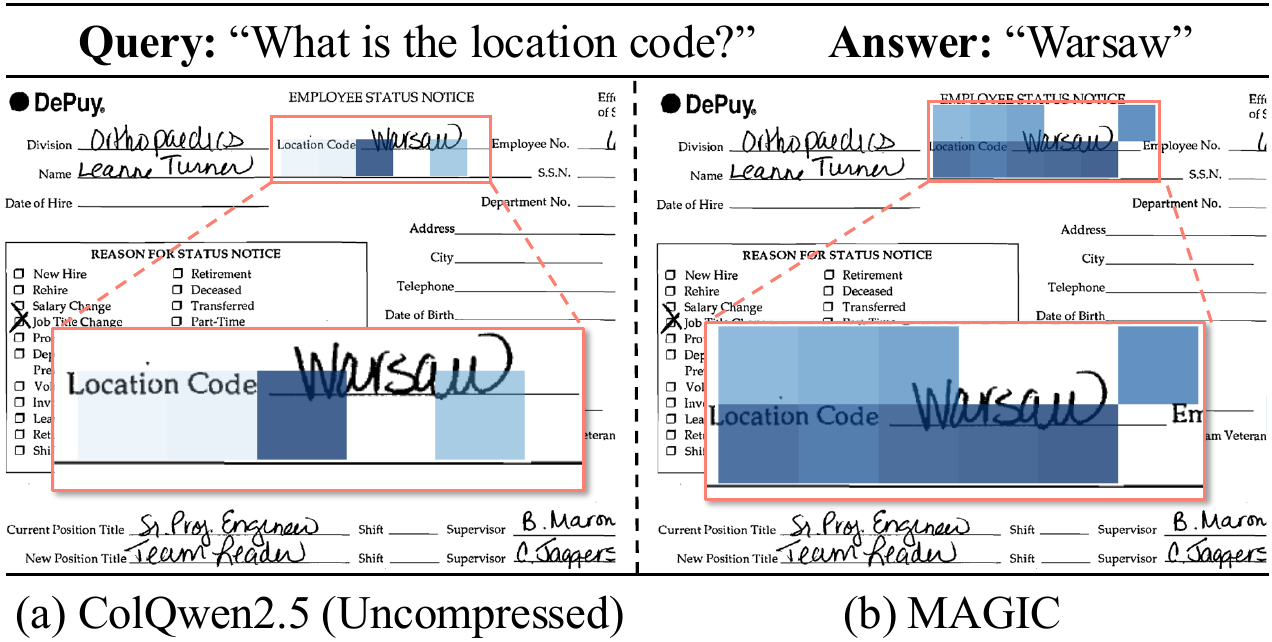}
    \captionof{figure}{
        Query-grounded evidence before and after MAGIC compression.
    }
    \label{fig:interpretability}
\end{minipage}
\par

\noindent \textbf{Storage and Latency.}
We finally assess storage, single-query latency, and full offline indexing time in Table~\ref{tab:storage-latency}.
Because all compressed indexes use the same online MaxSim scorer, storage and query latency mainly follow the retained-vector budget.
At \(r=0.01\), MAGIC reduces index size by about \(99\times\) and latency by about \(17\times\) relative to the uncompressed index.
Its offline indexing time remains close to the 112.6 ms/page encoder floor, requiring 122.8 ms/page at \(r=0.1\) and 119.9 ms/page at \(r=0.01\), substantially lower than Light-ColPali and DocPruner.
These results show that MAGIC preserves the lightweight deployment profile of post-hoc compression while adding little offline overhead beyond embedding extraction.
Detailed measurement protocols are provided in Appendix~\ref{app:storage-latency-details}.

\begin{table}[t]
\caption{Storage, single-query latency, and offline indexing time on ViDoRe v1 with ColQwen2.5.}
\label{tab:storage-latency}
\centering
\resizebox{\columnwidth}{!}{
\begin{tabular}{lcccccc}
\toprule
\multirow{2}{*}{\textbf{Method}} &
\multicolumn{3}{c}{\(\boldsymbol{r=0.1}\)} & \multicolumn{3}{c}{\(\boldsymbol{r=0.01}\)} \\
\cmidrule(lr){2-4}\cmidrule(lr){5-7}
& \shortstack{Size (MB/1K)} & \shortstack{Latency (ms/query)} & \shortstack{Comp. (ms/page)}
& \shortstack{Size (MB/1K)} & \shortstack{Latency (ms/query)} & \shortstack{Comp. (ms/page)} \\
\midrule
ColQwen2.5    & 380.67 & 0.98 & 112.62 & 380.67 & 0.98 & 112.62 \\
\midrule
DocPruner     & 43.38 & 0.18 & 279.00 & 9.69 & 0.08 & 279.00 \\
Light-ColPali & 37.79 & 0.16 & 210.41 & 3.54 & 0.06 & 212.72 \\
MAGIC         & 38.24 & 0.17 & \textbf{122.79} & 3.86 & 0.06 & \textbf{119.91} \\
\bottomrule
\end{tabular}
}
\vspace{-1em}
\end{table}

\section{Conclusion}
This paper revisited post-hoc compression for multi-vector visual document retrieval from the perspective of retrieval-aligned capacity allocation. 
We showed that generic merging can misallocate scarce retained vectors when
retrieval demand is sparse and non-uniform, and introduced MAGIC to align patch coverage with this demand while avoiding overly concentrated facet usage. 
Across compression budgets and retrieval settings, MAGIC improves the accuracy--efficiency trade-off most clearly when the retained-vector budget is tight. These results suggest that marginal design offers a simple and effective way to make frozen multi-vector retrievers more storage- and latency-efficient.


\bibliography{custom}
\bibliographystyle{iclr2027_conference}

\clearpage
\appendix





\section{Method Details}
\label{sec:app-method-details}

\subsection{Complete MAGIC Algorithm}
\label{app:magic-algorithm}

Algorithm~\ref{alg:magic} summarizes the complete MAGIC compression procedure.

\begin{algorithm}[H]
\caption{MAGIC Compression}
\label{alg:magic}
\begin{algorithmic}[1]
\REQUIRE Patch-embedding set \(D=\{d_j\}_{j=1}^{N}\), keep-ratio \(r\), calibration query-token pool \(\mathcal{Q}^{\mathrm{cal}}\), temperatures \(\tau,\varepsilon\), outer iterations \(T_{\mathrm{out}}\), Sinkhorn iterations \(T_{\mathrm{sk}}\), step size \(\eta\)
\ENSURE Compressed facets \(F^{\mathrm{out}}\)
\STATE Normalize all \(d_j\in D\)
\STATE \(K\leftarrow \min\{N,\max(1,\lceil rN\rceil)\}\)
\STATE Estimate \(\widehat{w}\) by Eq.~\ref{eq:selection-estimator}; set marginals \(a_j\leftarrow N\widehat{w}_j\) and \(b_k\leftarrow (\sum_{j=1}^{N} a_j)/K\) in Eq.~\ref{eq:magic-objective}
\STATE Initialize \(F\) with farthest-first seeds from \(D\)
\FOR{\(t=1,\ldots,T_{\mathrm{out}}\)}
    \STATE \(M_{kj}\leftarrow f_k^\top d_j/\varepsilon\)
    \STATE Compute \(T\) by \(T_{\mathrm{sk}}\) iterations of Eq.~\ref{eq:sinkhorn} using marginals \(a\) and \(b\)
    \STATE Row-normalize \(T\) to \(W\)
    \STATE Update \(F\) by Eq.~\ref{eq:barycenter-update}
\ENDFOR
\STATE Assign patches and form \(F^{\mathrm{out}}\) by Eq.~\ref{eq:lloyd-readout}
\RETURN \(F^{\mathrm{out}}\)
\end{algorithmic}
\end{algorithm}

\subsection{Retrieval Misalignment Diagnostics}
\label{app:diagnostics}

\paragraph{Diagnostic Query Generation.}
For the diagnostic analysis in Figure~\ref{fig:motivation}, we randomly sample
1,000 examples from the ViDoRe Benchmark collection.\footnote{
\url{https://huggingface.co/collections/vidore/vidore-benchmark}} Following the
synthetic-query protocol of Light-ColPali~\citep{ma2025towards}, we prompt
Qwen3-VL-8B-Instruct to generate five diverse page-conditioned queries for each
sampled page using the following prompt:
\begin{PromptStyleBox}{Diagnostic Query Generation Prompt}
\begin{PromptBody}
Given the screenshot of a document/poster, you are asked to generate five
question that can be answered by looking at the image. The questions should be
relevant to the content of the image. The questions should be unique and be of
varying question types. The output should be formatted as:\\
1. Question 1\\
2. Question 2\\
3. Question 3\\
4. Question 4\\
5. Question 5\\
\end{PromptBody}
\end{PromptStyleBox}

\paragraph{Retrieval-Demand Lorenz Curve.}
These generated queries are used only for the diagnostic analysis and are
separate from the held-out calibration query-token pool used by MAGIC. For each
sampled page, we encode the page into patch embeddings \(D\) and encode the
generated queries into query-token embeddings using the same frozen retriever.
Let \(\mathcal{Q}^{\mathrm{diag}}\) denote the resulting set of query-token
embeddings. We weight all diagnostic query tokens uniformly and first define
the patch selected by each diagnostic query token as
\begin{equation}
    j^*(q;D)
    =
    \operatorname*{arg\,max}_{1\le r\le N} q^\top d_r .
    \label{eq:diagnostic-selected-patch}
\end{equation}
The normalized patch-demand vector is then
\begin{equation}
    w_j
    =
    \frac{1}{|\mathcal{Q}^{\mathrm{diag}}|}
    \sum_{q\in\mathcal{Q}^{\mathrm{diag}}}
    \mathbf{1}\!\left[j=j^*(q;D)\right].
    \label{eq:diagnostic-demand}
\end{equation}
Thus \(w_j\) is the fraction of diagnostic query tokens whose MaxSim-selected
patch is \(d_j\). To construct the demand Lorenz curve in
Figure~\ref{fig:motivation}(a), we sort the demand values in descending order,
denoted \(w_{(1)}\ge\cdots\ge w_{(N)}\), and compute the cumulative demand mass
covered by the top \(\rho\) fraction of patches:
\begin{equation}
    L_{\mathrm{top}}(\rho)
    =
    \sum_{j=1}^{\lceil \rho N\rceil} w_{(j)} .
    \label{eq:diagnostic-lorenz}
\end{equation}
The point at \(\rho=0.2\) corresponds to the annotation in
Figure~\ref{fig:motivation}(a). A curve far above the uniform diagonal indicates
that a small fraction of patches accounts for a large fraction of MaxSim
selections.

\paragraph{Compression Diagnostics.}
Given compressed facets \(F\), retrieval-demand-weighted covering error is
\begin{equation}
    C_{\mathrm{err}}(D,F;w)
    =
    \sum_{j=1}^{N} w_j \min_{k}
    \left(1-d_j^\top f_k\right).
    \label{eq:diagnostic-covering-error}
\end{equation}
For each patch \(d_j\), the inner term is the cosine covering loss to its
nearest retained facet. The coefficient \(w_j\) is the fraction of diagnostic
query tokens for which \(d_j\) is the MaxSim-selected patch. Thus, an error on a
frequently selected patch contributes more to \(C_{\mathrm{err}}\) than the same
geometric error on a rarely selected patch. Lower values therefore indicate
better preservation of retrieval-relevant patch evidence.

To measure how retrieval demand is distributed over retained facets, let
\(\pi_F(j)=\operatorname*{arg\,max}_{k}d_j^\top f_k\) assign each patch to its
nearest retained facet and \(m_k=\sum_{j:\pi_F(j)=k}w_j\) be the
retrieval-demand mass assigned to facet \(k\). We define
\begin{equation}
    K_{\mathrm{eff}}(F;w)
    =
    \exp\!\left(-\sum_{k=1}^{K}m_k\log m_k\right),
    \label{eq:diagnostic-keff}
\end{equation}
with zero-mass terms omitted. Since \(\sum_k m_k=1\), this entropy effective
count satisfies \(1\le K_{\mathrm{eff}}\le K\): it equals one when all
retrieval demand is absorbed by a single facet and approaches \(K\) when
retrieval demand is evenly spread over the retained facets.

For a method and keep ratio, we compute \(C_{\mathrm{err}}\) and the raw
\(K_{\mathrm{eff}}\) for each diagnostic page and report their average over the
diagnostic page set. Figure~\ref{fig:motivation}(b) applies this diagnostic to
Light-ColPali and MAGIC. \(C_{\mathrm{err}}\) is plotted on the left axis, while
the raw \(K_{\mathrm{eff}}\) is plotted on the right axis against the allocated
facet budget \(K\), shown as a reference line. Larger \(C_{\mathrm{err}}\)
indicates poorer coverage of patches with high retrieval demand. A large gap
between \(K_{\mathrm{eff}}\) and \(K\) indicates that the compressed index uses
only a small effective number of retrieval-active facets despite being allocated
more retained vectors.

\subsection{Proof of Proposition 1}
\label{app:compression-bound-proof}

This section provides the complete proof of Proposition~1, which bounds the
expected MaxSim score degradation between the original patch embeddings and the
compressed facets. We first establish the single-token bound and then extend it
to multi-token MaxSim queries. 
Throughout this section, \(w\) denotes the population retrieval-demand
distribution in Eq.~\ref{eq:maxsim-source-marginal}, and \(\pi(j)\) denotes
the patch-to-facet assignment used by the compressed representation, so
\(f_{\pi(j)}\) is the compressed facet assigned to patch \(d_j\).

\paragraph{Single-token bound.}
For a fixed query token \(q\), define
\begin{equation}
    \begin{aligned}
    h_D(q)=\max_j q^\top d_j,
    \qquad
    h_F(q)=\max_k q^\top f_k .
    \end{aligned}
    \label{eq:app-single-token-scores}
\end{equation}
and let
\(\Delta_+(q;D,F)=[h_D(q)-h_F(q)]_+\) denote the positive single-token
MaxSim degradation, where \([x]_+=\max\{x,0\}\).
Let
\(j^*=j^*(q)=\operatorname*{arg\,max}_j q^\top d_j\). Because
\(f_{\pi(j^*)}\in F\), we have
\begin{equation}
    h_F(q)=\max_k q^\top f_k \ge q^\top f_{\pi(j^*)}.
    \label{eq:app-facet-lower-bound}
\end{equation}
Therefore,
\begin{equation}
\begin{aligned}
    \Delta_+(q;D,F)
    &=[h_D(q)-h_F(q)]_+ \\
    &\le [q^\top d_{j^*}-q^\top f_{\pi(j^*)}]_+ \\
    &\le |q^\top(d_{j^*}-f_{\pi(j^*)})| \\
    &\le \lVert q\rVert_2\lVert d_{j^*}-f_{\pi(j^*)}\rVert_2 \\
    &\le \lVert d_{j^*}-f_{\pi(j^*)}\rVert_2 .
\end{aligned}
    \label{eq:app-single-token-replacement-bound}
\end{equation}
The first inequality in Eq.~\ref{eq:app-single-token-replacement-bound} uses
Eq.~\ref{eq:app-facet-lower-bound}. The second uses \([x]_+\le |x|\), the
third is Cauchy--Schwarz, and the last uses \(\lVert q\rVert_2\le1\).
Taking expectation over \(q\sim\mathcal{P}_Q\) and partitioning by the event
\(j^*(q)=j\) gives
\begin{equation}
\begin{aligned}
    \mathbb{E}_q[\Delta_+(q;D,F)]
    &\le
    \sum_{j=1}^{N}\Pr[j^*(q)=j]\lVert d_j-f_{\pi(j)}\rVert_2 \\
    &=
    \sum_{j=1}^{N}w_j\lVert d_j-f_{\pi(j)}\rVert_2 .
\end{aligned}
    \label{eq:app-single-token-demand-bound}
\end{equation}
This partition groups query tokens by their MaxSim-selected patch. On the event
\(j^*(q)=j\), the replacement loss in the bound is
\(\lVert d_j-f_{\pi(j)}\rVert_2\), and the event probability is exactly the
retrieval demand \(w_j\).
By the weighted root-mean-square inequality,
\begin{equation}
    \sum_{j=1}^{N}w_j\lVert d_j-f_{\pi(j)}\rVert_2
    \le
    \sqrt{\sum_{j=1}^{N}w_j\lVert d_j-f_{\pi(j)}\rVert_2^2}.
    \label{eq:app-rms-bound}
\end{equation}
This step applies because \(w\) is a normalized nonnegative distribution. Since
all vectors are unit-normalized,
\begin{equation}
    \lVert d_j-f_{\pi(j)}\rVert_2^2
    =
    2(1-d_j^\top f_{\pi(j)}).
    \label{eq:app-unit-distance}
\end{equation}
Combining Eqs.~\ref{eq:app-single-token-demand-bound}--\ref{eq:app-unit-distance}
gives
\begin{equation}
\begin{aligned}
    \mathbb{E}_q[\Delta_+(q;D,F)]
    &\le
    \sum_{j=1}^{N}w_j\lVert d_j-f_{\pi(j)}\rVert_2 \\
    &\le
    \sqrt{
    2\sum_{j=1}^{N}w_j
    \left(1-d_j^\top f_{\pi(j)}\right)
    },
\end{aligned}
    \label{eq:app-single-token-weighted-bound}
\end{equation}
which restates Eq.~\ref{eq:weighted-bound}.

\paragraph{Multi-token MaxSim extension.}
The single-token bound applies to one term in the late-interaction score; a full
MaxSim query score sums such token-level maxima over all query tokens, as in
Eq.~\ref{eq:maxsim}. Let \(Q=\{q_i\}_{i=1}^{M}\sim
\mathcal{P}_{\mathrm{query}}\) denote a multi-token query instance with fixed
nonnegative token weights \(\alpha_i\), and let \(A=\sum_i\alpha_i>0\). The
weights \(\alpha_i\) allow a general weighted MaxSim score; the standard score in
Eq.~\ref{eq:maxsim} is recovered by setting \(\alpha_i=1\) for all query
tokens. Define
\begin{equation}
    \begin{aligned}
    S_D(Q)=\sum_{i=1}^{M}\alpha_i h_D(q_i),
    \qquad
    S_F(Q)=\sum_{i=1}^{M}\alpha_i h_F(q_i).
    \end{aligned}
    \label{eq:app-multitoken-scores}
\end{equation}
Here, \(S_D(Q)\) is the weighted MaxSim score computed with the original patch
embeddings, and \(S_F(Q)\) is the corresponding score computed with the
compressed facets. We define
\(\Delta_+(Q;D,F)=[S_D(Q)-S_F(Q)]_+\) as the positive multi-token MaxSim score
degradation.
Expectations in this paragraph are taken over
\(Q\sim\mathcal{P}_{\mathrm{query}}\). The aggregate MaxSim selection mass and
its normalized distribution are
\begin{equation}
    \begin{aligned}
    W_j=
    \mathbb{E}_Q\!\left[
    \sum_{i=1}^{M}\alpha_i
    \mathbf{1}\{j^*(q_i)=j\}
    \right],
    \qquad
    \bar{w}_j=\frac{W_j}{A}.
    \end{aligned}
    \label{eq:app-multitoken-selection-mass}
\end{equation}
Because \(\sum_j W_j=A\), \(\bar{w}\) is a normalized patch distribution.
Equivalently, \(\bar{w}\) is the token-level selection distribution obtained by
first sampling \(Q\sim\mathcal{P}_{\mathrm{query}}\), then sampling token
\(q_i\) with probability \(\alpha_i/A\). Thus \(\bar{w}\) plays the same role
for multi-token scores as \(w\) does for \(q\sim\mathcal{P}_Q\) in
Eq.~\ref{eq:maxsim-source-marginal}.
Since \([\sum_i x_i]_+\le\sum_i[x_i]_+\) and \(\alpha_i\ge0\), we have
\begin{equation}
\begin{aligned}
    \mathbb{E}_Q[\Delta_+(Q;D,F)]
    &\le
    \mathbb{E}_Q\!\left[
    \sum_{i=1}^{M}\alpha_i
    [h_D(q_i)-h_F(q_i)]_+
    \right] \\
    &\le
    \mathbb{E}_Q\!\left[
    \sum_{i=1}^{M}\alpha_i
    \lVert d_{j^*(q_i)}-f_{\pi(j^*(q_i))}\rVert_2
    \right] \\
    &=
    \mathbb{E}_Q\!\left[
    \begin{aligned}
    &\sum_{i=1}^{M}\alpha_i
    \sum_{j=1}^{N}\mathbf{1}\{j^*(q_i)=j\} \\
    &\qquad\cdot
    \lVert d_j-f_{\pi(j)}\rVert_2
    \end{aligned}
    \right] \\
    &=
    \sum_{j=1}^{N}W_j\lVert d_j-f_{\pi(j)}\rVert_2 \\
    &=
    A\sum_{j=1}^{N}\bar{w}_j\lVert d_j-f_{\pi(j)}\rVert_2 \\
    &\le
    A\sqrt{
    2\sum_{j=1}^{N}\bar{w}_j
    \left(1-d_j^\top f_{\pi(j)}\right)
    }.
\end{aligned}
    \label{eq:app-multitoken-bound}
\end{equation}
The second inequality in Eq.~\ref{eq:app-multitoken-bound} applies the
single-token bound in Eq.~\ref{eq:app-single-token-replacement-bound} to each
query token. The following equality partitions each token by the event
\(j^*(q_i)=j\) using indicator variables. The next equality exchanges the
finite sums with the expectation, yielding the aggregate selection mass \(W_j\)
defined in Eq.~\ref{eq:app-multitoken-selection-mass}.
For standard MaxSim, \(\alpha_i=1\), \(A=M\), and \(\bar{w}_j\) is the
expected fraction of query tokens in a sampled query that select patch \(j\).
If the token-level distribution \(\mathcal{P}_Q\) in
Eq.~\ref{eq:maxsim-source-marginal} is induced by first sampling
\(Q\sim\mathcal{P}_{\mathrm{query}}\) and then sampling one of its query tokens
uniformly, then \(\bar{w}=w\). Therefore, for the full multi-token MaxSim score
in Eq.~\ref{eq:maxsim},
\begin{equation}
    \begin{aligned}
    \mathbb{E}_{Q\sim\mathcal{P}_{\mathrm{query}}}
    [\Delta_+(Q;D,F)]
    &\le
    M\sqrt{
    2\sum_{j=1}^{N}w_j
    \left(1-d_j^\top f_{\pi(j)}\right)
    } \\
    &=
    M\sqrt{2\mathcal{L}_{\mathrm{RD}}(F,\pi)} .
    \end{aligned}
    \label{eq:app-full-query-rd-bound}
\end{equation}
Thus reducing the retrieval-demand-weighted surrogate in
Eq.~\ref{eq:rd-surrogate} tightens an upper bound on the expected positive
degradation of the full multi-token MaxSim score by
Eq.~\ref{eq:app-full-query-rd-bound}.

\subsection{Smoothed Retrieval-Demand Estimator}
\label{app:retrieval-demand-estimator}

The smoothed estimator in Eq.~\ref{eq:selection-estimator} approximates the
population retrieval-demand distribution \(w\), which is unknown at indexing
time. This distribution defines the retrieval-demand surrogate in
Eq.~\ref{eq:rd-surrogate} and the source marginal used in the OT objective in
Eq.~\ref{eq:magic-objective}. MAGIC estimates it from the held-out calibration
query-token pool \(\mathcal{Q}^{\mathrm{cal}}\) using a soft approximation to
hard MaxSim selection.

For \(\tau>0\), define
\begin{equation}
    \begin{aligned}
    X_j(q;\tau)
    &=
    \frac{\exp(q^\top d_j/\tau)}
    {\sum_{\ell=1}^{N}\exp(q^\top d_{\ell}/\tau)}, \\
    w_j^{(\tau)}
    &=
    \mathbb{E}_{q\sim\mathcal{P}_Q}[X_j(q;\tau)] .
    \end{aligned}
    \label{eq:app-smoothed-demand}
\end{equation}
The empirical estimator used by MAGIC is
\begin{equation}
    \widehat{w}_j(\mathcal{Q},\tau)
    =
    \frac{1}{|\mathcal{Q}|}
    \sum_{q\in\mathcal{Q}}X_j(q;\tau).
    \label{eq:app-smoothed-demand-estimator}
\end{equation}
Eq.~\ref{eq:app-smoothed-demand-estimator} specializes to
Eq.~\ref{eq:selection-estimator} when
\(\mathcal{Q}=\mathcal{Q}^{\mathrm{cal}}\).
For a query token with a unique MaxSim maximizer, \(j^*(q)\), the softmax
score in Eq.~\ref{eq:app-smoothed-demand} concentrates on that maximizer:
\begin{equation}
    \lim_{\tau\to0}X_j(q;\tau)
    =
    \mathbf{1}\{j=j^*(q)\}.
    \label{eq:app-softmax-hard-limit}
\end{equation}
Therefore,
\begin{equation}
    \lim_{\tau\to0}w_j^{(\tau)}
    =
    \Pr_{q\sim\mathcal{P}_Q}[j^*(q)=j]
    =
    w_j .
    \label{eq:app-smoothed-demand-limit}
\end{equation}
The quality of \(\widehat{w}\) depends on how well
\(\mathcal{Q}^{\mathrm{cal}}\) represents the query-token distribution
\(\mathcal{P}_Q\): a more representative calibration pool yields a closer
approximation to the population retrieval demand. Increasing the calibration
pool size generally improves coverage of \(\mathcal{P}_Q\), but the downstream
retrieval metric need not improve monotonically once the main query-token modes
are covered. This is consistent with
Figure~\ref{fig:calibration-query-prior}(b), where performance improves from
small pools and then saturates; larger pools provide limited additional benefit
and may introduce redundant or less relevant calibration signal.

Eq.~\ref{eq:app-smoothed-demand-limit} shows that the estimator recovers the
hard MaxSim selection distribution as \(\tau\to0\). In practice, MAGIC uses a
finite temperature instead of hard argmax counts because the softmax estimate is
less sensitive to near-ties and local score noise while still preserving the
hard-selection limit.

\section{Experimental Protocols}
\label{sec:app-experimental-details}

\begin{table*}[t]
\caption{Public sources for backbone checkpoints and baseline implementations used in the experiments.}
\label{tab:app-backbones-baselines}
\centering
\resizebox{\columnwidth}{!}{
\begin{tabular}{ll}
\toprule
\textbf{Model} & \textbf{Public source} \\
\midrule
\multicolumn{2}{l}{\textbf{Backbones}} \\
ColPali~\citep{faysse2024colpali} & \url{https://huggingface.co/vidore/colpali-v1.3} \\
ColQwen2.5~\citep{faysse2024colpali} & \url{https://huggingface.co/vidore/colqwen2.5-v0.2} \\
Jina-v4~\citep{gunther2025jina} & \url{https://huggingface.co/jinaai/jina-embeddings-v4} \\
ColNomic~\citep{nomicembedmultimodal2025} & \url{https://huggingface.co/nomic-ai/colnomic-embed-multimodal-3b} \\
\midrule
\multicolumn{2}{l}{\textbf{Post-hoc compression baselines}} \\
DocPruner~\citep{yan2025docpruner} & \url{https://arxiv.org/abs/2509.23883} \\
Voronoi Pruning~\citep{kankanampati2026voronoi} & \url{https://github.com/yash-reddy/voronoi-pruning} \\
Light-ColPali~\citep{ma2025towards} & \url{https://arxiv.org/abs/2506.04997} \\
ColChunk~\citep{yan2026visual} & \url{https://arxiv.org/abs/2604.10167} \\
Prune-then-Merge~\citep{yan2026sculpting} & \url{https://arxiv.org/abs/2602.19549} \\
\midrule
\multicolumn{2}{l}{\textbf{Learned compact multi-vector baselines}} \\
AGC~\citep{qin2026multi} & \url{https://huggingface.co/hltcoe/AGC_qwen2.5-vl_colpali} \\
MetaEmbed~\citep{xiao2025metaembed} & \url{https://github.com/facebookresearch/MetaEmbed} \\
\midrule
\multicolumn{2}{l}{\textbf{Single-vector retrieval baselines}} \\
VLM2Vec~\citep{jiang2025vlmvec} & \url{https://huggingface.co/TIGER-Lab/VLM2Vec-Qwen2VL-2B} \\
VLM2Vec-V2~\citep{meng2026vlmvecv} & \url{https://huggingface.co/VLM2Vec/VLM2Vec-V2.0} \\
VisRAG~\citep{yu2025visrag} & \url{https://huggingface.co/openbmb/VisRAG-Ret} \\
Jina-v4~\citep{gunther2025jina} & \url{https://huggingface.co/jinaai/jina-embeddings-v4} \\
UniME-V2~\citep{gu2026unime} & \url{https://huggingface.co/TianchengGu/UniME-V2-Qwen2VL-2B} \\
Qwen3-VL-Embedding-2B~\citep{li2026qwen3} & \url{https://huggingface.co/Qwen/Qwen3-VL-Embedding-2B} \\
Qwen3-VL-Embedding-8B~\citep{li2026qwen3} & \url{https://huggingface.co/Qwen/Qwen3-VL-Embedding-8B} \\
\bottomrule
\end{tabular}
}
\end{table*}

\subsection{Datasets and Evaluation Protocol}
\label{app:datasets-evaluation}

\paragraph{Benchmarks.}
Following Light-ColPali~\citep{ma2025towards}, the primary evaluation uses six subsets from ViDoRe v1~\citep{faysse2024colpali}: ArxivQA (ArxivQ)~\citep{li2024multimodal}, DocVQA (DocQ)~\citep{mathew2021docvqa}, InfoVQA (InfoQ)~\citep{mathew2022infographicvqa}, TAT-DQA (TATQ)~\citep{zhu2022towards}, TabFQuAD (TabF), and Shift Project (Shift). 
Additional synthetic ViDoRe v1 subsets are omitted because they offer little
headroom for distinguishing modern visual document retrievers whose performance
is already near saturation.

ViDoRe v2~\citep{mace2025vidore} is used to test multilingual transfer under
more realistic retrieval conditions. Compared with ViDoRe v1, it reduces
extractive-query bias, includes long-form and cross-document queries, and uses
hybrid synthetic and human-reviewed query construction. The evaluation includes
three multilingual topic-specific subsets, covering ESG reports, biomedical
lectures, and economics reports, with queries in English, French, German, and
Spanish, together with a fully human-labeled ESG subset.

ViDoSeek~\citep{wang2025vidorag} is used for the additional reasoning-intensive
evaluation in Appendix~\ref{app:generalization}. It targets RAG over visually rich document
collections, with about 1,200 questions spanning text, charts, tables, complex
layouts, and both single-hop and multi-hop reasoning. Each query is associated
with a unique answer and specific reference pages, making it suitable for
evaluating page-level retrieval in large document collections.

\paragraph{Evaluation Protocol and Metrics.}
Each page image is treated as one retrieval candidate. For post-hoc compression
experiments, all methods use the same frozen retriever and are evaluated with
the standard MaxSim score over their retained vectors. The main metrics are
nDCG@5 and Recall@5, reported as percentages. All aggregate ViDoRe v1 results,
including Table~\ref{tab:posthoc-main}, the budget-sweep curves, and the
ablation studies, report macro-averages over the six selected subsets.
Budget-sweep experiments use keep ratios
\(r\in\{0.5,0.25,0.1,0.05,0.025,0.01\}\) unless a baseline only supports a
subset of these operating points.

\subsection{Backbones and Baseline Implementations}
\label{app:backbones-baselines}

\paragraph{Backbones.}
The public checkpoints used for frozen embedding extraction are listed in the
top block of Table~\ref{tab:app-backbones-baselines}. ColQwen2.5 is the default
retriever for the main post-hoc compression experiments, while ColPali, Jina-v4,
and ColNomic are used for backbone-transfer comparisons.

\paragraph{Baselines.}
The main comparison evaluates post-hoc compressors that operate on frozen
multi-vector page embeddings. This group includes simple baselines implemented
directly, namely 1D-Pooling, 2D-Pooling, and K-Means, as well as prior post-hoc
compression methods: DocPruner~\citep{yan2025docpruner}, Voronoi
Pruning~\citep{kankanampati2026voronoi}, Light-ColPali~\citep{ma2025towards},
ColChunk~\citep{yan2026visual}, and
Prune-then-Merge~\citep{yan2026sculpting}.

For the simple baselines, 1D-Pooling partitions the patch-embedding sequence
into non-overlapping windows and averages the embeddings in each window. Its
pooling factor \(\{2,4,10,20,40,100\}\) realizes keep ratios
\(\{0.5,0.25,0.1,0.05,0.025,0.01\}\). K-Means uses spherical Lloyd clustering
on L2-normalized patch embeddings, assigns patches by cosine similarity, and
represents each cluster by its normalized mean. Its merge factor uses the same
set \(\{2,4,10,20,40,100\}\) to match the same keep ratios, and the number of
Lloyd iterations is fixed to 10. 2D-Pooling arranges patch embeddings on their
spatial grid and applies average pooling with a square \(m\times m\) kernel.
Because this construction changes the vector count by roughly \(m^2\), it does
not match every one-dimensional keep ratio exactly. Accordingly, 2D-Pooling is
evaluated only at the two main operating points: \(r\approx0.1\) with \(m=3\),
and \(r=0.01\) with \(m=10\).

The appendix further includes two complementary comparisons. AGC~\citep{qin2026multi}
and MetaEmbed~\citep{xiao2025metaembed} are learned compact multi-vector
retrievers and are evaluated under a fixed document-vector budget. Single-vector
retrievers, including VLM2Vec~\citep{jiang2025vlmvec},
VLM2Vec-V2~\citep{meng2026vlmvecv}, VisRAG~\citep{yu2025visrag},
Jina-v4~\citep{gunther2025jina}, UniME-V2~\citep{gu2026unime}, and
Qwen3-VL-Embedding~\citep{li2026qwen3}, are evaluated under matched
embedding-size budgets. Public implementations and checkpoints are listed in
Table~\ref{tab:app-backbones-baselines}. Methods without public checkpoints or
code are implemented following their corresponding papers.

\subsection{Calibration Protocol}
\label{app:calibration-protocol}

The calibration query-token pool \(\mathcal{Q}^{\mathrm{cal}}\) is constructed
from \texttt{vidore/colpali\_train\_set}\footnote{\url{https://huggingface.co/datasets/vidore/colpali_train_set}},
the training set used by ColPali and ColQwen2.5. This same training source is used to build
\(\mathcal{Q}^{\mathrm{cal}}\) for all evaluation datasets, keeping calibration
queries disjoint from the evaluation queries in ViDoRe v1, ViDoRe v2, and
ViDoSeek.

The visual activation and diversity criterion follows the construction in
Eq.~\ref{eq:score-biased-fps}. In all experiments, the corpus visual
dictionary contains \(P=200\) diverse patch representatives, and the final
calibration pool contains \(L_{\mathrm{cal}}=1000\) query-token embeddings. A
separate \(\mathcal{Q}^{\mathrm{cal}}\) is built once for each frozen backbone,
because query-token and patch embeddings lie in backbone-specific embedding
spaces and cannot be shared across backbones.

For the calibration-pool ablation in
Figure~\ref{fig:calibration-query-prior}, all variants use the same
held-out training-query source and the same default pool size unless the calibration
token count is explicitly varied. The \textit{w/o VA} variant removes the
visual activation score and selects query tokens using only the diversity
criterion. The \textit{w/o Div} variant removes the diversity step and keeps
the query tokens with the highest visual activation scores. The Random variant
samples query tokens uniformly. In the calibration token-count ablation, only
\(L_{\mathrm{cal}}\) is changed, while the visual dictionary size and other
MAGIC hyperparameters remain fixed.

These ablations support the main observation that visual grounding and token
diversity are complementary. In Figure~\ref{fig:calibration-query-prior}(a), MAGIC performs
best overall, while Random remains competitive because uniform sampling provides
implicit diversity over the training-query pool.
The gain over Random is small at \(r=0.1\) and becomes
clearer at the tighter budget \(r=0.05\), where visual grounding helps allocate
the limited retained-vector capacity toward query-relevant visual content.
In Figure~\ref{fig:calibration-query-prior}(b), the calibration token-count sweep shows rapid saturation, with little
change beyond \(L_{\mathrm{cal}}=1000\), suggesting that the retrieval-demand
estimate does not require a large calibration pool once visual activation and diversity are enforced.
Since \(L_{\mathrm{cal}}\) counts query-token embeddings rather than
full queries, the default value \(L_{\mathrm{cal}}=1000\) corresponds to only a
few dozen full queries in practice.

\subsection{MAGIC Hyperparameters}
\label{app:magic-hyperparameters}

Table~\ref{tab:app-magic-hyperparameters} reports the default hyperparameters
used by MAGIC in the main experiments. Unless otherwise specified, the same
values are used across datasets, backbones, and keep ratios.

\begin{table}[t]
\caption{Default MAGIC hyperparameters used in the main experiments.}
\label{tab:app-magic-hyperparameters}
\centering
\begin{tabular}{lll}
\toprule
\textbf{Hyperparameter} & \textbf{Value} & \textbf{Meaning} \\
\midrule
\(L_{\mathrm{cal}}\) & 1000 & calibration query-token count \\
\(\tau\) & 0.05 & selection temperature \\
\(\varepsilon\) & 0.05 & entropic temperature \\
\(T_{\mathrm{out}}\) & 5 & outer iterations \\
\(T_{\mathrm{sk}}\) & 5 & Sinkhorn iterations \\
\(\eta\) & 0.982 & barycenter step size \\
\bottomrule
\end{tabular}
\end{table}

\subsection{Interpretability Visualization}
\label{app:interpretability-details}

This section provides the visualization protocol for the interpretability
experiment in Figure~\ref{fig:interpretability}, where MAGIC is evaluated at
keep ratio \(r=0.05\). 
For the uncompressed ColQwen2.5 visualization, each query token selects the
highest-scoring original document patch under MaxSim, and the selected patches
are overlaid on the page image. For MAGIC, each query token first selects a
compressed facet under the same MaxSim rule,
\(k^*(q)=\operatorname*{arg\,max}_{k}q^\top f_k^{\mathrm{out}}\). The selected
facet is then back-projected to original image patches using the learned
transport/assignment weights: the contribution of patch \(d_j\) is given by
\(T_{k^*(q)j}\). Source patches are ranked by this assignment mass, and
high-mass patches are highlighted to compare the compressed representation with
the uncompressed patch-level evidence in the original image coordinate system. Because each compressed facet can aggregate multiple original patches,
the back-projected MAGIC visualization may highlight more patches than the
uncompressed MaxSim visualization.

\subsection{Storage and Latency Measurement}
\label{app:storage-latency-details}

We report the efficiency results in Table~\ref{tab:storage-latency} on ViDoRe v1
with ColQwen2.5. All measurements use the same 2,000-page corpus sampled from
ViDoRe v1, with 743.5 document vectors per page on average before compression.
Index size is reported per 1K pages and is computed from each method's measured
mean retained-vector count, using fp32 128-dimensional vectors and decimal MB
units. 
DocPruner retains the 11 ColQwen2.5 prompt tokens in addition to selected image patches, which explains its larger index size.
Query latency is measured with batch size 1 by exhaustive MaxSim against
the full-corpus index on one NVIDIA H20 GPU. The latency is averaged over 1,000
distinct queries, each replayed five times.

The reported compression time is the full offline indexing cost per page,
including encoder forward computation and post-hoc compression, averaged over
all 2,000 pages and five runs. The standard ColQwen2.5 embedding-extraction
path runs with Flash Attention and outputs projected 128-dimensional patch
embeddings, which gives an encoder floor of 112.6 ms/page. Light-ColPali and
MAGIC operate on these projected patch embeddings and add their post-hoc
compression cost on top of this floor. DocPruner requires last-layer attention
scores for patch pruning and therefore uses Eager Attention, which raises its
encoder cost to 279.0 ms/page. 
For MAGIC, the reported time includes per-page retrieval-demand estimation, OT-based compression, and spherical-Lloyd readout.

\begin{table*}[t]
\caption{Comparison with learned efficient baselines on ViDoRe v2 using a fixed budget of 64 document embeddings per page. The metric is nDCG@5 (\%). ``EN'' is the English language subset, ``Multi. Avg.'' denotes the average over multilingual queries, and ``Human'' denotes the human-labeled ESG subset.}
\label{tab:learned-main}
\centering
\resizebox{\columnwidth}{!}{
\begin{tabular}{lc|cc|cc|ccc}
\toprule
\multirow{2}{*}{\textbf{Method}} & \multirow{2}{*}{\shortstack{\textbf{Vectors}\\ \textbf{/ page}}} &
\multicolumn{2}{c|}{\textbf{Biomedical Lectures}} &
\multicolumn{2}{c|}{\textbf{Economics Reports}} &
\multicolumn{3}{c}{\textbf{ESG Reports}} \\
\cmidrule(lr){3-4}\cmidrule(lr){5-6}\cmidrule(lr){7-9}
& & EN & Multi. Avg. & EN & Multi. Avg. & EN & Multi. Avg. & Human \\
\midrule
ColQwen2.5     & 759   & 63.2 & 60.3 & 61.0 & 57.2 & 60.6 & 56.2 & 64.6 \\
\midrule
AGC            &  64   & 60.2 & 57.7 & 62.8 & 56.3 & 54.5 & 53.1 & 54.5 \\
MetaEmbed      &  64   & 61.7 & 58.7 & 62.3 & 55.5 & 62.6 & 57.4 & 63.7 \\
\rowcolor{gray!20} \textbf{MAGIC} &  64   & 60.6 & 57.8 & 60.3 & 56.8 & 57.2 & 53.2 & 55.4 \\
\bottomrule
\end{tabular}
}
\end{table*}

\section{Extended Results}
\label{sec:app-additional-results}

\subsection{Generalization on ColNomic and ViDoSeek}
\label{app:generalization}

\begin{figure}[t]
    \centering
    \makebox[\columnwidth][c]{%
        \begin{minipage}[t]{0.35\columnwidth}
            \centering
            \includegraphics[width=\linewidth]{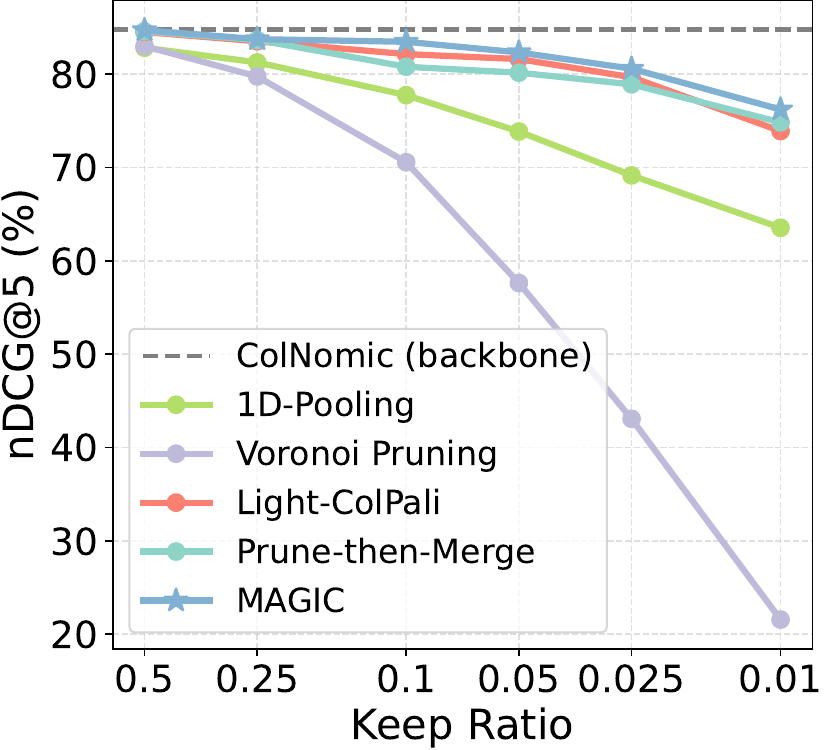}\\
            \small (a) ViDoRe v1 - ColNomic
        \end{minipage}%
        \hspace{0.05\columnwidth}%
        \begin{minipage}[t]{0.35\columnwidth}
            \centering
            \includegraphics[width=\linewidth]{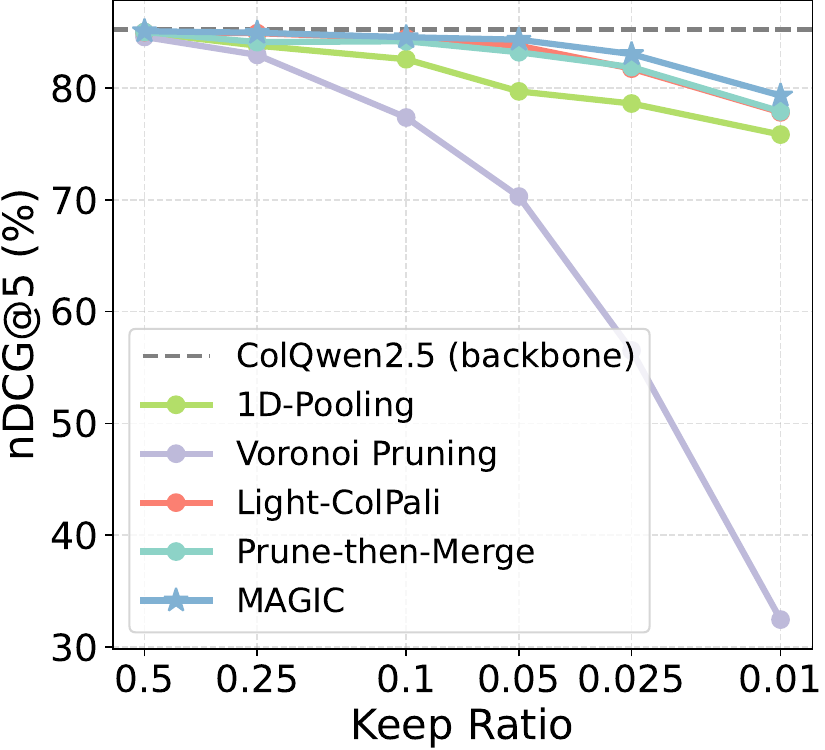}\\
            \small (b) ViDoSeek - ColQwen2.5
        \end{minipage}%
    }
    \caption{Additional budget-sweep retrieval performance on ColNomic and
    ViDoSeek. Each curve reports nDCG@5 over the keep ratio.}
    \label{fig:app-additional-budget-curves}
\end{figure}

Figure~\ref{fig:app-additional-budget-curves} extends the budget-sweep
evaluation to an additional retrieval backbone and an additional benchmark.
On ViDoRe v1 with ColNomic, MAGIC remains the strongest method across the
compression range. The curves are close at mild compression, but the gap widens
as the keep ratio decreases: at \(r=0.01\), MAGIC remains around the mid-70s
nDCG@5, while 1D-Pooling drops much more sharply and Voronoi Pruning collapses
under the tight budget.

The same trend holds on ViDoSeek with ColQwen2.5. MAGIC consistently stays
closest to the uncompressed backbone curve and is most clearly separated from
generic pooling and pruning baselines at the smallest keep ratios. These
additional results complement the generalization results in Figure~\ref{fig:dense-budget-curves}: the
retrieval-demand source marginal and balanced target marginal continue to
provide effective capacity allocation beyond the default backbone and ViDoRe
evaluation setting.

\subsection{Comparison with Learned Baselines}
\label{app:learned-baselines}

This comparison evaluates MAGIC against learned efficient baselines under a
fixed embedding-count budget. Learned methods such as
MetaEmbed~\citep{xiao2025metaembed} and AGC~\citep{qin2026multi} expose capacity
through a learned number of retained embeddings rather than through a post-hoc
keep ratio over the original page patches. A shared budget of 64 document
embeddings per page is used to test whether training-free post-hoc compression
remains competitive with methods that learn compact representations under the
same absolute document-vector budget. All methods in this comparison use
Qwen2.5-VL-3B as the underlying VLM.

Table~\ref{tab:learned-main} uses the uncompressed ColQwen2.5 retriever as a
reference point. It keeps 759 document vectors per page on average, whereas the
compressed methods are evaluated at 64 vectors per page. ColQwen2.5, AGC, and
MAGIC share the same underlying training-data setting: AGC is trained with the
same data as ColQwen2.5, and MAGIC compresses the frozen ColQwen2.5 page
embeddings without any additional training. Under this matched data setting,
MAGIC is competitive with the trained AGC compressor and achieves higher
nDCG@5 on six of the seven ViDoRe v2 columns, with the largest gains on the ESG
subsets. 
This comparison shows that MAGIC's marginal-guided post-hoc compression can
match or exceed a trained compact compressor in this setting without additional
training.

MetaEmbed is trained with substantially more data, making the comparison to
MetaEmbed not strictly one-to-one: the ColQwen2.5/AGC setting uses about
one-twentieth of MetaEmbed's training scale. This larger training scale is
reflected in the table, where MetaEmbed even surpasses the uncompressed
ColQwen2.5 reference on the English Economics subset and the English and
multilingual-average ESG subsets. This training-scale difference should be
considered when interpreting MAGIC's lower scores than MetaEmbed on several
subsets, while the comparison still shows that a training-free post-hoc
compressor can remain competitive with learned compact retrievers under the
same 64-vector budget.

\begin{table*}[t]
\caption{Comparison with single-vector baselines under matched document-side
embedding-size budgets on ViDoRe v1. The uncompressed ColQwen2.5 retriever is treated as a reference point.
MAGIC uses 12 or 16 retained 128-dimensional facets, corresponding to 1536 and 2048 dimensions, respectively.
Single-vector baselines use one global page embedding at the
corresponding dimensionality. The 2048-dimensional group also includes VisRAG,
whose closest available setting uses a 2304-dimensional global embedding. The
metric is nDCG@5 (\%).}
\label{tab:app-single-vector-baselines}
\centering
\resizebox{\columnwidth}{!}{
\begin{tabular}{llccccccc}
\toprule
\textbf{Method} & \textbf{Representation} & \textbf{ArxivQ} & \textbf{DocQ} & \textbf{InfoQ} & \textbf{TabF} & \textbf{TATQ} & \textbf{Shift} & \textbf{Avg.} \\
\midrule
ColQwen2.5 & \(763\times128\) vectors & 88.65 & 62.60 & 91.25 & 89.67 & 81.40 & 87.87 & 83.57 \\
\midrule
\multicolumn{9}{c}{\textbf{Embedding-size budget = 1536 dimensions}} \\
VLM2Vec & \(1\times1536\) vector & 60.88 & 32.68 & 74.30 & 82.10 & 31.22 & 65.16 & 57.72 \\
VLM2Vec-V2 & \(1\times1536\) vector & 79.79 & 44.96 & 84.41 & \textbf{91.23} & 48.33 & 70.87 & 69.93 \\
UniME-V2 & \(1\times1536\) vector & 63.79 & 37.39 & 73.06 & 82.31 & 40.59 & 53.86 & 58.50 \\
\rowcolor{gray!20}  \textbf{MAGIC} & \(12\times128\) facets & \textbf{84.36} & \textbf{51.26} & \textbf{85.03} & 86.76 & \textbf{67.85} & \textbf{74.78} & \textbf{75.01} \\
\midrule
\multicolumn{9}{c}{\textbf{Embedding-size budget \(\approx\) 2048 dimensions}} \\
VisRAG  & \(1\times2304\) vector & 80.65 & 43.21 & 84.90 & 76.80 & 50.25 & 61.54 & 66.23 \\
Jina-v4 & \(1\times2048\) vector & 84.74 & 51.58 & 87.05 & \textbf{94.99} & 65.30 & 81.64 & 77.55 \\
Qwen3-VL-Embedding-2B & \(1\times2048\) vector & 74.97 & 42.61 & 85.43 & 93.97 & 57.76 & 80.66 & 72.56 \\
Qwen3-VL-Embedding-8B & \(1\times2048\) vector & 84.05 & 47.22 & \textbf{88.92} & 94.04 & 66.08 & \textbf{82.24} & 77.09 \\
\rowcolor{gray!20} \textbf{MAGIC} & \(16\times128\) facets & \textbf{85.37} & \textbf{55.66} & 88.32 & 87.85 & \textbf{73.12} & 76.15 & \textbf{77.74} \\
\bottomrule
\end{tabular}
}
\end{table*}

\begin{figure*}[t]
    \centering
    \begin{minipage}[t]{0.32\textwidth}
        \centering
        \includegraphics[width=\linewidth]{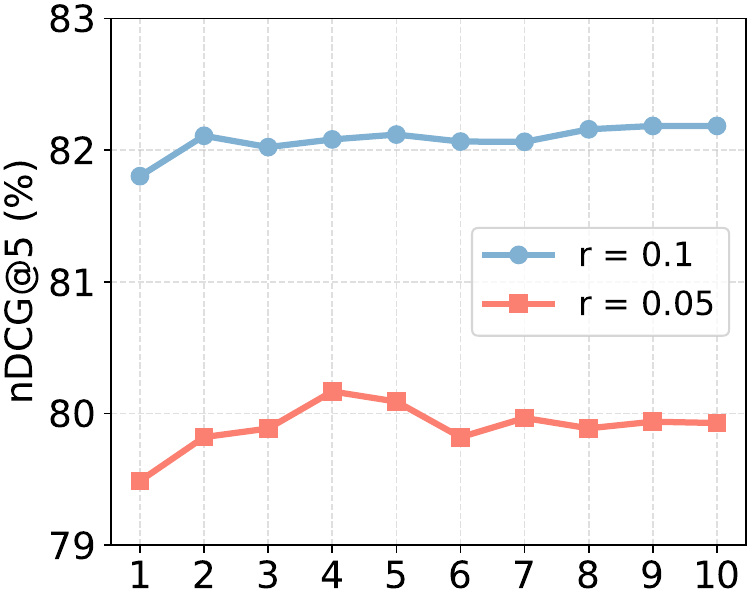}\\
        \small (a) Sinkhorn iterations \(T_{\mathrm{sk}}\)
    \end{minipage}\hfill
    \begin{minipage}[t]{0.32\textwidth}
        \centering
        \includegraphics[width=\linewidth]{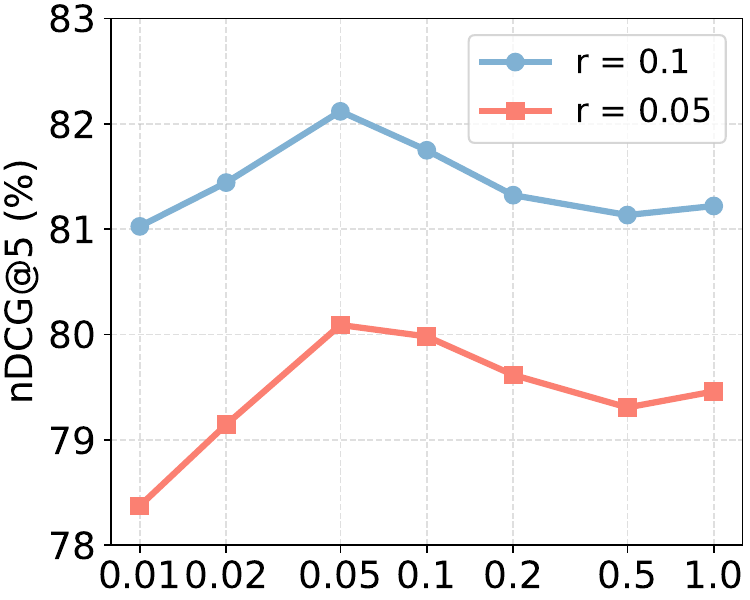}\\
        \small (b) Selection temperature \(\tau\)
    \end{minipage}\hfill
    \begin{minipage}[t]{0.32\textwidth}
        \centering
        \includegraphics[width=\linewidth]{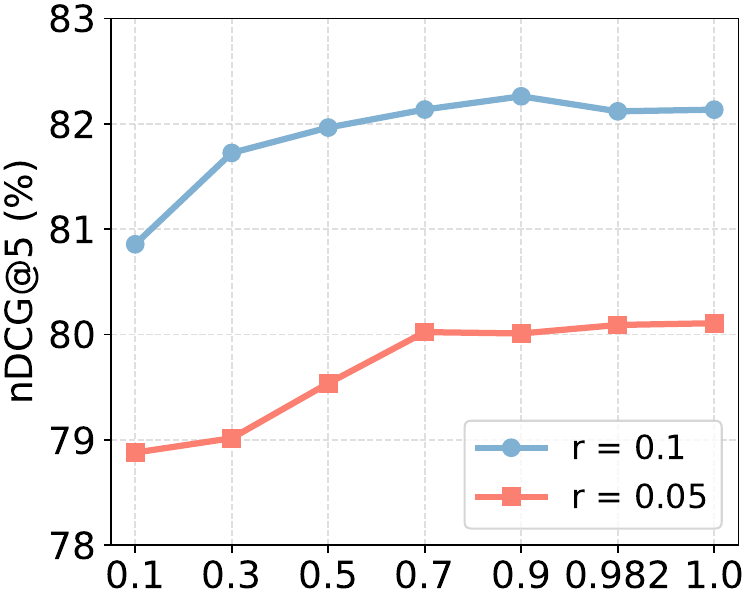}\\
        \small (c) Barycenter step size \(\eta\)
    \end{minipage}
    \caption{Additional MAGIC hyperparameter sensitivity. The curves report
    nDCG@5 across hyperparameter values at two keep ratios.}
    \label{fig:app-additional-hyperparameters}
\end{figure*}

\subsection{Comparison with Single-Vector Baselines}
\label{app:single-vector-baselines}

Single-vector VDR baselines provide a useful reference for the storage--accuracy
trade-off studied in this paper. These methods encode each page into one global
embedding and support standard approximate nearest-neighbor search without
MaxSim scoring. They therefore provide a compact retrieval setting against which
MAGIC's compressed multi-vector representations can be compared.

The comparison uses matched document-side embedding-size budgets rather than
matched vector counts, because the embedding dimensionality differs across
single-vector retrievers. Table~\ref{tab:app-single-vector-baselines} includes
the uncompressed ColQwen2.5 retriever as a reference point and reports two
budget groups. In the 1536-dimensional group, MAGIC keeps 12 128-dimensional
facets per page. In the 2048-dimensional group, MAGIC keeps 16 128-dimensional
facets per page; VisRAG is included in this group because its closest available
setting uses a 2304-dimensional global embedding.

The single-vector baselines include
VLM2Vec~\citep{jiang2025vlmvec}, VLM2Vec-V2~\citep{meng2026vlmvecv},
UniME-V2~\citep{gu2026unime}, VisRAG~\citep{yu2025visrag},
Jina-v4~\citep{gunther2025jina}, and Qwen3-VL-Embedding~\citep{li2026qwen3}.
This set covers visual document retrievers, universal multimodal embedding
models, and recent VLM-based embedding systems, giving a broad comparison
against representative single-vector retrieval approaches. These models differ
in their training data and training recipes, so the comparison also reflects
backbone and data differences beyond the document-side embedding budget.

The results show that compressed multi-vector retrieval remains competitive
under comparable storage budgets. At 1536 dimensions, MAGIC achieves the best
average nDCG@5, improving over the strongest single-vector baseline,
VLM2Vec-V2, by 5.08 points. The gains are especially large on DocQ, TATQ, and
Shift, while VLM2Vec-V2 remains stronger on TabF. At the approximate
2048-dimensional budget, the gap becomes smaller because Jina-v4 and
Qwen3-VL-Embedding-8B are strong single-vector retrievers. MAGIC still obtains
the highest average score, with clear gains on DocQ and TATQ, while Jina-v4
performs best on TabF and Qwen3-VL-Embedding-8B performs best on InfoQ and
Shift. The ColQwen2.5 reference row helps contextualize these task-specific
advantages: on TabF, VLM2Vec-V2, Jina-v4, and both Qwen3-VL-Embedding variants
surpass the uncompressed ColQwen2.5 retriever, suggesting that differences in
training data or model design can be substantial on some subsets. These results
indicate that retaining a small set of MaxSim-compatible facets can preserve
fine-grained matching behavior that is difficult to capture with a single global
page embedding, especially under tighter embedding-size budgets.

\subsection{Additional Hyperparameter Analysis}
\label{app:additional-hyperparameters}

Figure~\ref{fig:app-additional-hyperparameters} complements the
optimization analysis in Figure~\ref{fig:optimization-choices}. Increasing the
number of Sinkhorn iterations improves performance from one iteration and then
quickly saturates; beyond two to five iterations, the curves remain nearly
flat. This supports the default \(T_{\mathrm{sk}}=5\), which is sufficient for
stable transport plans without making the inner solver expensive.

The selection temperature \(\tau\) is more sensitive. Very small values make
the estimator close to hard MaxSim counts, as discussed in
Eq.~\ref{eq:app-smoothed-demand-limit}. This can make the estimated source
marginal overly sparse and sensitive to near-ties or local score noise, so
performance drops, especially at the tighter budget. Moderate smoothing around
\(\tau=0.05\) gives the best results by assigning most mass to the strongest
patches while still retaining information from close competitors. Larger
values make the estimated source marginal too diffuse. The barycenter step-size
sweep covers \(\eta\in[0.1,1.0]\). Small step sizes under-update the facets
toward their demand-weighted barycenters and therefore give lower retrieval
accuracy. Performance improves as \(\eta\) increases and becomes nearly flat
after about \(\eta=0.7\), with the default \(\eta=0.982\) lying in this
stable high-performing region. Overall, these ablations show that MAGIC does
not require delicate tuning for the inner OT solver, while the retrieval-demand
smoothing temperature is the most important hyperparameter among the three.

\subsection{Additional Qualitative Results}
\label{app:additional-qualitative}

\begin{figure*}[p]
    \centering
    \includegraphics[width=\linewidth]{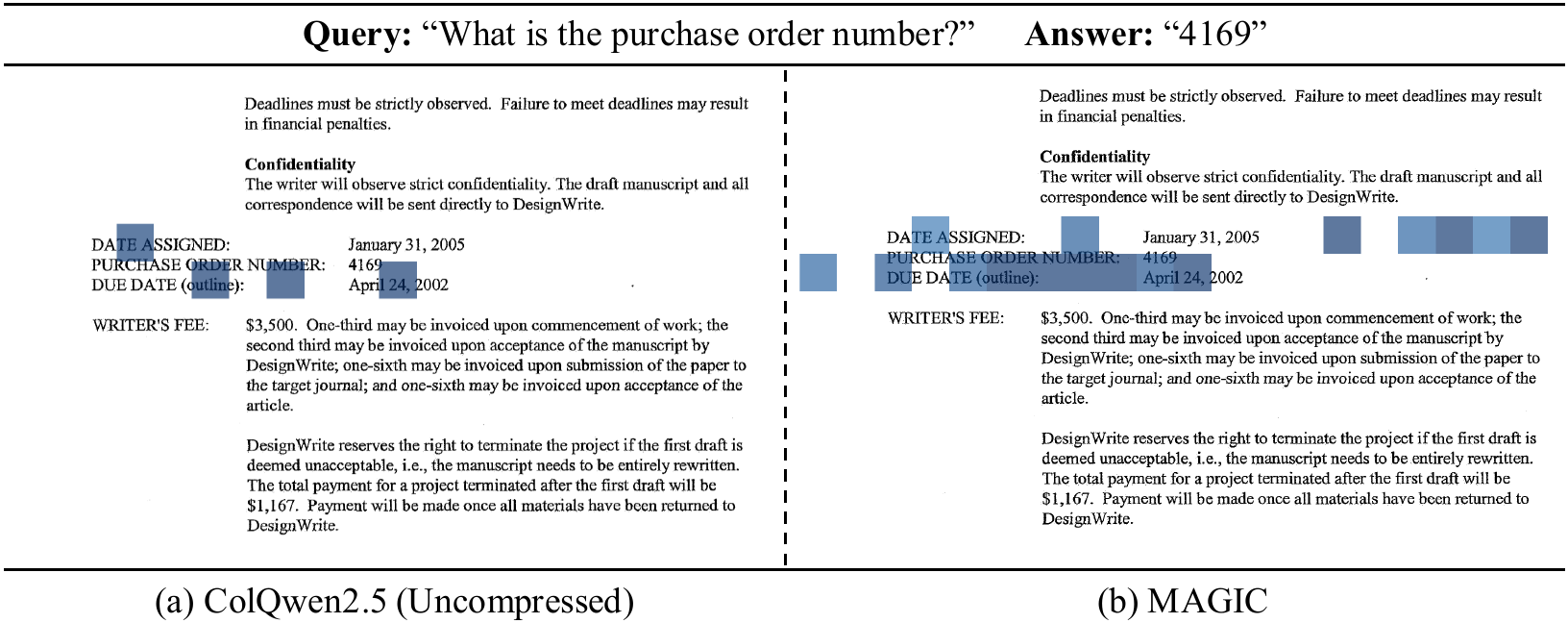}
    \caption{Additional qualitative example on DocVQA. The query asks for a
    purchase-order number, and MAGIC preserves the answer-bearing field while
    operating at a 0.05 keep ratio.}
    \label{fig:app-qualitative-docvqa}

    \vspace{1em}
    \includegraphics[width=\linewidth]{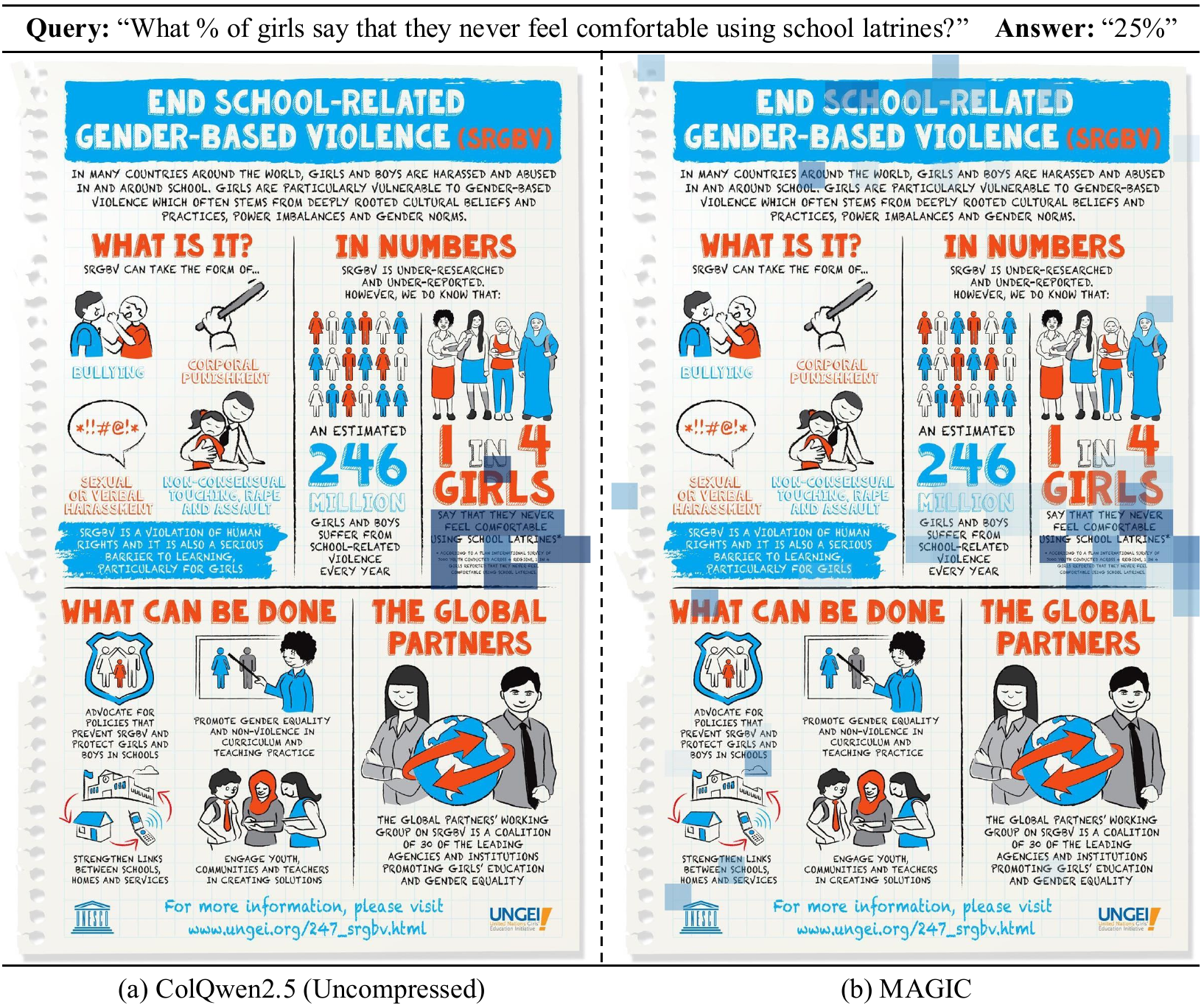}
    \caption{Additional qualitative example on InfoVQA. MAGIC retains evidence
    around the infographic region supporting the queried percentage.}
    \label{fig:app-qualitative-infovqa}
\end{figure*}

\begin{figure*}[t]
    \centering
    \includegraphics[width=\linewidth]{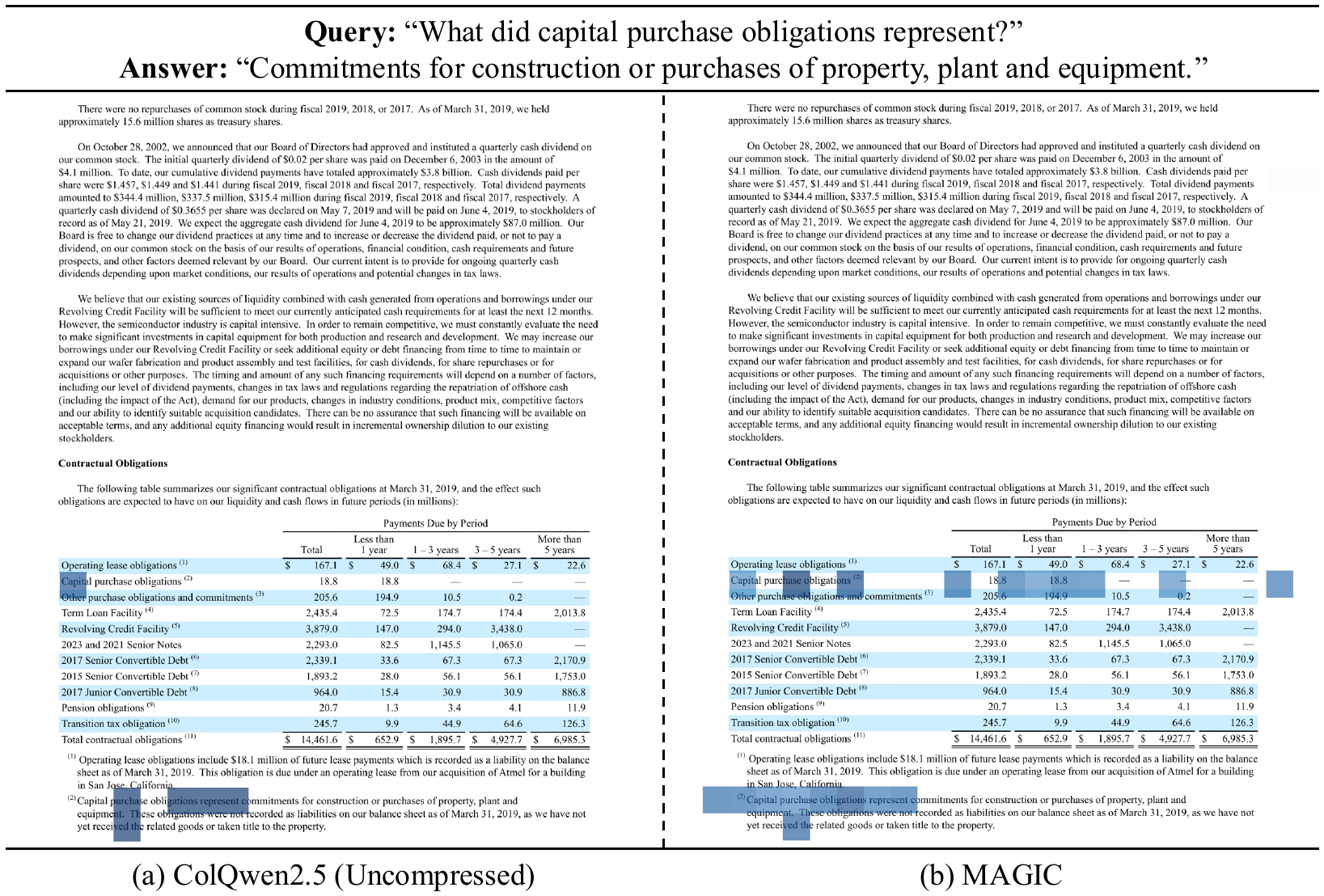}
    \caption{Additional qualitative example on TAT-DQA. MAGIC keeps the
    definition evidence needed to answer the financial-document query.}
    \label{fig:app-qualitative-tatdqa}
\end{figure*}

\begin{figure*}[t]
    \centering
    \includegraphics[width=\linewidth]{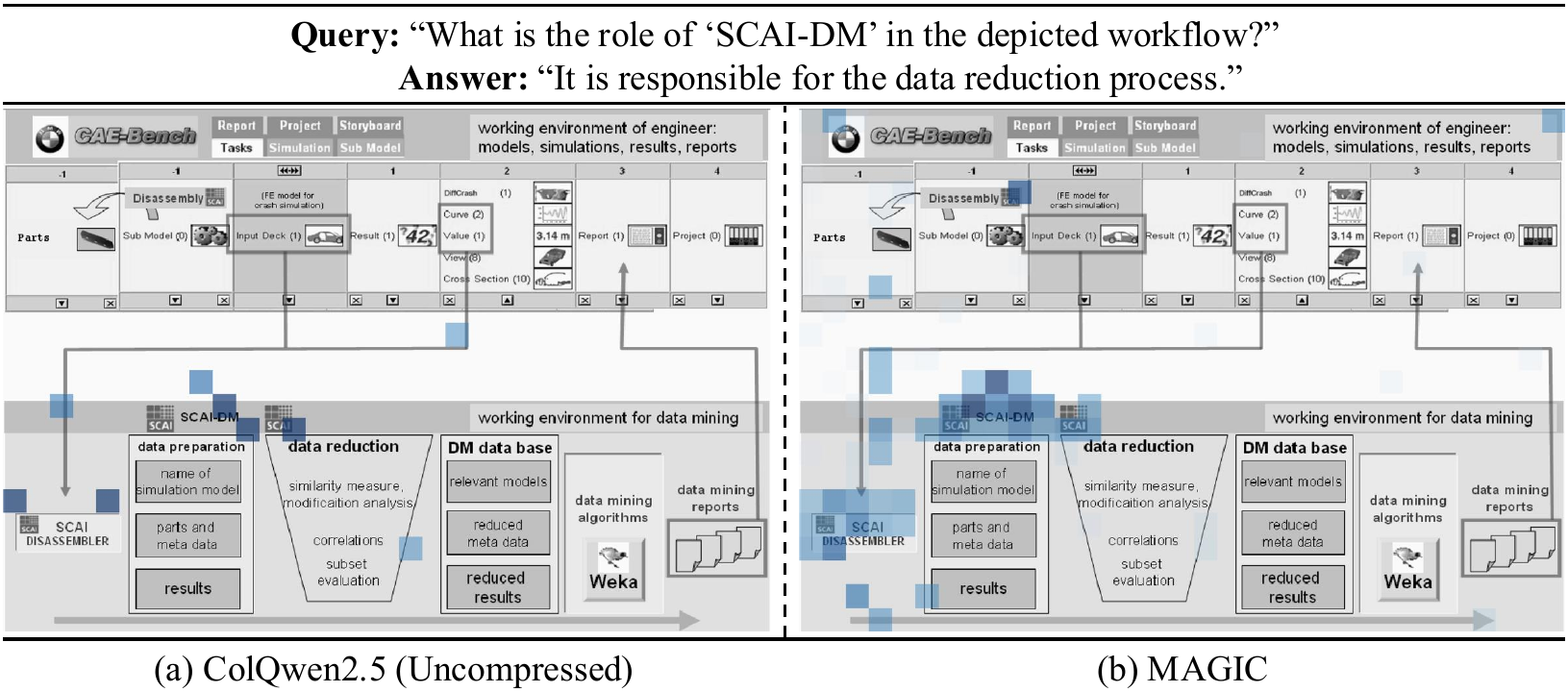}
    \caption{Additional qualitative example on ArxivQA. MAGIC preserves the
    workflow evidence around SCAI-DM.}
    \label{fig:app-qualitative-arxivqa}
\end{figure*}

Figures~\ref{fig:app-qualitative-docvqa}--\ref{fig:app-qualitative-arxivqa}
provide additional qualitative examples for the interpretability experiment in
Figure~\ref{fig:interpretability}. Each example follows the same visualization protocol as
Appendix~\ref{app:interpretability-details}: the uncompressed ColQwen2.5 patch
view is compared with MAGIC at a 0.05 keep ratio, and highlighted regions mark
the image evidence receiving the largest MaxSim assignment mass.

The examples cover four visually different document subsets from ViDoRe v1. 
In DocVQA, MAGIC preserves the local form field containing the purchase-order 
identifier, showing that small but answer-critical numbers are retained after 
compression. In InfoVQA, the highlighted evidence remains concentrated on the 
infographic region that supports the queried percentage, indicating that MAGIC 
can preserve localized layout evidence in dense visual pages. In TAT-DQA, the
compressed facets continue to emphasize the textual definition associated with
capital purchase obligations, which is the evidence needed for the financial
reasoning query. In ArxivQA, MAGIC highlights the workflow component around
SCAI-DM, showing that the compressed representation can also preserve 
diagram-level semantic evidence. Because each compressed
facet can represent multiple source patches, MAGIC may highlight a broader set
of patches than the uncompressed view, but the highlighted regions remain
centered on the answer-bearing evidence. These qualitative results support the
interpretability finding that MAGIC preserves MaxSim-compatible evidence across
forms, infographics, financial documents, and scientific diagrams.

\section{Complexity and Limitations}
\label{sec:app-discussion}

\subsection{MAGIC Complexity}
\label{app:complexity}

\paragraph{MAGIC cost.}
For \(N\) patches, \(K\) retained facets, embedding dimension \(p\),
\(L_{\mathrm{cal}}\) calibration query-token embeddings, \(T_{\mathrm{out}}\)
outer iterations, and \(T_{\mathrm{sk}}\) Sinkhorn iterations, the per-page
offline time complexity of MAGIC is
\begin{equation}
    T_{\mathrm{MAGIC}}
    =
    \mathcal{O}\!\left(L_{\mathrm{cal}}Np\right)
    +
    \mathcal{O}\!\left(T_{\mathrm{out}}KNp\right)
    +
    \mathcal{O}\!\left(T_{\mathrm{out}}T_{\mathrm{sk}}KN\right).
    \label{eq:complexity-time}
\end{equation}
The first term in Eq.~\ref{eq:complexity-time} estimates the
retrieval-demand source marginal in Eq.~\ref{eq:selection-estimator}. The
second term covers the facet--patch similarity computation and barycenter
updates across outer iterations, and the third term covers log-domain Sinkhorn
updates. The deterministic initialization and final spherical-Lloyd readout
each cost \(\mathcal{O}(KNp)\), which is absorbed by the second term for
\(T_{\mathrm{out}}\ge1\). In the experiments, \(p=128\), \(L_{\mathrm{cal}}\),
\(T_{\mathrm{out}}\), and \(T_{\mathrm{sk}}\) are fixed constants. The main
scaling variables are therefore the patch count \(N\) and the retained-vector
budget \(K=rN\), where \(0<r\le1\). Under this setting,
Eq.~\ref{eq:complexity-time} can be read
as approximately \(\mathcal{O}(L_{\mathrm{cal}}Np+rN^2p)\), with the
\(K\times N\) facet--patch computation dominating the optimization loop when
the keep ratio \(r\) is small. The factors \(p\) and \(L_{\mathrm{cal}}\) are
kept in this simplified expression because they determine the wall-clock cost
of the dot-product computations: \(p\) is the embedding dimension of each
similarity evaluation, and \(L_{\mathrm{cal}}\) controls the retrieval-demand
estimation cost, which does not shrink with \(K\).

The per-page working memory is \(\mathcal{O}(Np+Kp+KN)\) for the patch
embeddings, facets, and \(K\times N\) transport matrix, plus the shared
calibration pool \(\mathcal{O}(L_{\mathrm{cal}}p)\), which is reused across
pages for a fixed backbone. With fixed \(p\), the per-page memory is
dominated by storing the original page vectors and the \(K\times N\) transport
matrix.

\paragraph{Comparison with Light-ColPali.}
Light-ColPali is the most relevant complexity comparison because it is a
representative and competitive post-hoc merging baseline in the main
experiments and, like MAGIC, compresses frozen page embeddings without
retraining the retriever. The
\(K\times N\) structure gives MAGIC a practical complexity advantage in the
aggressive-compression regime. Light-ColPali applies Ward hierarchical
agglomerative clustering to the \(N\) page vectors, which requires constructing
a full pairwise/hierarchical structure. Its per-page post-hoc cost is therefore
\(\mathcal{O}(N^2p+N^2)\) time and \(\mathcal{O}(N^2+Np)\) memory. This cost
is largely independent of the final retained-vector budget \(K\), whereas MAGIC optimizes a
budget-dependent \(K\times N\) transport problem after estimating retrieval
demand. When \(K=rN\) with small keep ratio \(0<r\le1\), the dominant MAGIC
optimization term scales with \(rN^2p\) rather than the full \(N^2p\)
pairwise vector comparison underlying hierarchical clustering. This
budget-dependent computation is consistent with
Table~\ref{tab:storage-latency}, where MAGIC has lower offline computation time
than Light-ColPali at both reported keep ratios. Online retrieval uses the same
late-interaction scorer as the original retriever, with the document-side
vector count reduced from \(N\) to \(K\).

\subsection{Limitations}
\label{app:limitations}

The main limitations of MAGIC concern calibration coverage, extreme compression
budgets, and backbone-specific preprocessing.

(1) MAGIC relies on a held-out calibration query-token pool to estimate
retrieval-demand source marginals. Although the calibration pool is small and
disjoint from evaluation queries, its quality depends on how well it represents
the target retrieval workload. If rare query intents or domain-specific visual
patterns are underrepresented, the estimated demand may underweight patches
needed for those queries.

(2) Under extremely aggressive compression, multiple fine-grained patches must
be represented by a single facet. Small text fields, dense tables, or
layout-specific evidence may therefore still be lost.

(3) The calibration pool and compressed index are backbone-specific because
different retrievers define different embedding spaces. Deploying MAGIC across
multiple backbones therefore requires separate offline calibration and
compression.

\end{document}